\documentclass[5p,final,authoryear,twocolumn]{elsarticle}
\usepackage{soul,framed} 
\usepackage{subcaption}

\usepackage[cmex10]{amsmath}
\usepackage{amssymb}
\usepackage{url}
\usepackage{color, colortbl}
\usepackage{booktabs}
\usepackage[utf8]{inputenc}
\newcommand{\revision}[1]{#1}

\begin{document}
\begin{frontmatter}
	\title{Discriminative and Generative Models for Anatomical Shape Analysis \\ on Point Clouds with Deep Neural Networks}

\author{Benjam{\'\i}n Guti{\'e}rrez-Becker\corref{cor3}}
\author{Ignacio Sarasua\corref{cor3}} 
\author{Christian Wachinger\corref{cor1}}
\author{\\ \small{for the Alzheimer's Disease Neuroimaging Initiative}\corref{cor2}} 
\cortext[cor3]{Joint first authors.}
\cortext[cor1]{Corresponding Author. Address: Waltherstr. 23, 81369 München, Germany; Email: christian.wachinger@med.uni-muenchen.de}
\cortext[cor2]{Data used in preparation of this article were obtained from the Alzheimer's Disease Neuroimaging Initiative (ADNI) database (adni.loni.usc.edu). As such, the investigators within the ADNI contributed to the design and implementation of ADNI and/or provided data but did not participate in analysis or writing of this report.
A complete listing of ADNI investigators can be found at:
\url{http://adni.loni.usc.edu/wp-content/uploads/how_to_apply/ADNI_Acknowledgement_List.pdf}}
\address{
Lab for Artificial Intelligence in Medical Imaging (AI-Med),\\ Department of Child and Adolescent Psychiatry, Psychosomatics and Psychotherapy,
University Hospital, LMU Munich}

	
	\begin{abstract}
		
		We introduce deep neural networks for the analysis of anatomical shapes that learn a low-dimensional shape representation from the given task, instead of relying on hand-engineered representations. Our framework is modular and consists of several computing blocks that perform fundamental shape processing tasks. 
		The networks operate on unordered point clouds and provide invariance to similarity transformations, avoiding the need to identify point correspondences between shapes.
		Based on the framework, we assemble a discriminative model for disease classification and age regression, as well as a generative model for the accruate reconstruction of shapes. 
		In particular, we propose a conditional generative model, where the condition vector provides a mechanism to control the generative process.
		For instance, it enables to assess shape variations specific to a particular diagnosis, when passing it as side information. 
		Next to working on single shapes, we introduce an extension for the joint analysis of multiple anatomical structures, where the simultaneous modeling of multiple structures can lead to a more compact encoding and a better understanding of disorders. 
		We demonstrate the advantages of our framework in comprehensive experiments on real and synthetic data. 
		The key insights are that (i) learning a shape representation specific to the given task yields higher performance than alternative shape descriptors, (ii) multi-structure analysis is both more efficient and more accurate than single-structure analysis, and (iii)  point clouds generated by our model capture morphological differences associated to Alzheimer’s disease, to the point that they can be used to train a discriminative model for disease classification. 
		Our framework naturally scales to the analysis of large datasets, giving it the potential to learn characteristic variations in large populations.
		%
		%
		%
		%
		%
	\end{abstract}
	
	
	\end{frontmatter}

	\section{Introduction}

Over the last decades, a variety of shape analysis techniques  have been developed for modeling the human anatomy  from medical images \citep{ng2014shape}.
These methods have become a mainstay in medical image analysis, not only because of their utility in providing priors for segmentation, but also because of their value in quantifying shape changes between subjects and populations~\citep{shen2012detecting}. Shape analysis helps in localizing anatomical changes, which can yield a better understanding of morphological changes due to aging and disease~\citep{gerardin2009multidimensional,wachinger2017latent}. 

Given that the morphology of organs across a population is highly heterogeneous, modeling and quantifying these shape variations is a challenging task. Thanks to the growing availability of large-scale medical imaging datasets, we have now the possibility to model these underlying shape variations in the population more accurately.
Unfortunately, working on large sample sizes comes with computational challenges, which can limit the practical application of traditional methods for shape analysis \citep{ng2014shape}. 
In addition, imaging datasets  usually come with valuable phenotypic information of the patient. 
This large amount of available data, paired with recent advances in machine learning, calls for the development of a data-driven and learning-based  shape analysis framework that can benefit from the large amount of image data and provides a mechanism to include prior information in the analysis. 

Many fields in medical image analysis have recently been revolutionized by the introduction of  deep neural networks~\citep{litjens2017survey}. These approaches have the ability to learn complex, hierarchical feature representations that have proven to outperform hand-crafted features in a variety of medical imaging applications.  
\revision{Also in shape analysis, learning a shape representation may offer advantages in contrast to working with pre-defined parameterizations like  point distribution models \citep{Cootes1995}, spectral signatures~\citep{Wachinger2015,reuter2006laplace}, spherical harmonics \citep{gerardin2009multidimensional}, medial representations \citep{Gorczowski2007}, and diffeomorphisms \citep{miller2014diffeomorphometry,pennec2019riemannian}.} 
One of the main reasons for the success of neural networks in image analysis is the use of convolutional layers, which take advantage of the shift-invariance properties of images \citep{Bronstein2017}. However, the use of deep networks in medical shape analysis is still largely unexplored; mainly because  typical shape representations such as point clouds and meshes do not possess an underlying Euclidean or grid-like structure. 


In this work, we introduce a modular and versatile framework for shape analysis with deep neural networks. 
The framework consists of computing blocks for processing shapes that can be assembled to perform a variety of tasks in shape analysis. 
First, we construct a discriminative model for the prediction of Alzheimer's disease and age, where the network learns a shape descriptor that is optimal for the given task. 
Second, we propose a generative model for the unsupervised learning of shape variations within a population. In particular, we introduce a \emph{conditional} generative model to have the ability to also include non-image data. 
Our results show that the neural network learns modes of variation that capture complex shape changes, yielding a compact representation and the generation of realistic samples.

Our framework is based on a deep neural network architecture, which operates directly on a point cloud representation of organs. 
Point clouds present a raw, lightweight and simple parameterization that avoid complexities involved with meshes and that is trivial to obtain given a segmented surface. 
\revision{While point clouds do not cover mesh connectivity, they offer high flexibility as there is no need of pre-defining a topology or aligning all  shapes to a template, e.g., \cite{ranjan2018generating}.
}
Our framework offers the following advantages: 1) it is invariant to similarity transformations, avoiding the need to pre-align the shapes to be analyzed; 2) it is invariant to the ordering of the elements in the point cloud, meaning that computing  correspondences between points across shapes is not necessary; 3) it does not impose any constraints on the topology of the shapes, providing high flexibility; 4) it scales to analyzing large shape datasets and therefore has the potential to learn characteristic variations in large populations.

Finally, our framework cannot only process a single shape but simultaneously multiple shapes.
In many applications, we do not only want to study one organ in isolation, as several anatomical structures can be segmented from an image. 
To provide a more holistic picture of the anatomy, we therefore want to jointly process multiple shapes. 
For discriminative tasks, this can increase the classification accuracy. 
For generative tasks, this will give us access to a joint low-dimensional representation that captures anatomical variations of multiple structures.

	\subsection{Related Work}
	\def\real{{\mathbb{R}}}
	\def\ourmethod{{OURMETHOD}}
	
	Previous work in medical shape analysis can be roughly divided in either discriminative or generative approaches. 
	Discriminative approaches aim at finding shape descriptors, which allow to quantify and characterize the shape of anatomical structures. These shape descriptors are then used to perform either classification or regression tasks. Thickness and volume measurements of brain structures have been used to perform age regression \citep{becker2018gaussian} and to classify between healthy controls (HC) and patients \citep{costafreda2011automated}. Medial descriptors were used to perform discriminative analysis between HCs and autistic subjects \citep{Gorczowski2007}. Approaches based on spectral signatures have been previously  used to perform disease prediction and age regression~\citep{Wachinger2015}.  
	Different to these approaches, our proposed model is not based on the computation of  pre-defined shape features, but rather aims at find low-dimensional representations optimized for a particular task.
	
	Different to discriminative approaches, generative shape models are able not only to quantify shape variations, but also allow to generate new valid shapes similar to those the model observed during training.   The most common approach to produce generative models for medical shape analysis is based on Point Distribution Models (PDMs) \citep{cootes1992active}. PDMs represent shapes using point cloud representations, which consist of points distributed across the surface. A generative model based on PDM's is usually built by finding a mean shape and the principal modes of shape variation through the use of Principal Component Analysis (PCA). A challenge of PDM's is however that point correspondences have to be found between all shapes in the dataset. 
	This usually involves a registration step, which is not only challenging but also computationally expensive for large databases. Moreover, homologous features may not exist when comparing shapes that are subject to strong variations, e.g.,  over the course of brain development. While our method is also based on point clouds, we do not require correspondences between shapes.
	\revision{A variational autoencoder has been proposed for learning a low-dimensional shape representation from meshes~\citep{Shakeri2016}, where correspondences between meshes were computed with spectral matching~\citep{lombaert2013diffeomorphic}.} 	An autoencoder for shape analysis on binary images was proposed in~\citep{evan2019convolutional}. 
	Other popular generative shape models are based on skeletal representations \citep{pizer2013nested}, spherical harmonics \citep{gerardin2009multidimensional}, and \revision{deformations \citep{miller2014diffeomorphometry, durrleman2014morphometry,pennec2019riemannian}}.

	Conditional variational autoencoders (CVAE)~\citep{kingma2013auto,Sohn2015} are an extension of the generative model in variational autoencoders by introducing a condition vector, which allows to include prior information in the autoencoder. A CVAE has recently been used in medical imaging for 3D fetal skull reconstruction from 2D ultrasound \citep{cerrolaza20183d}. Conditional generative models have also recently become popular in the context of generative adversarial networks \citep{goodfellow2014generative}.  A conditional adversarial networks was proposed as a general-purpose solution to image-to-image translation problems \citep{isola2017image}. 
	In contrast to those previous work, we are proposing a conditional generative model for shape analysis on point cloud representations.
	
	An earlier version of this work has been presented at conferences~\citep{gutierrez2018deep,gutierrez2019learning}, and has been conceptualized and expanded in this article.

	\section{Method}\label{sec:method}
	
	We present a modular and versatile framework for shape analysis on point clouds with deep neural networks. 
	In section~\ref{sec:compblocks}, we introduce several computational blocks for processing point clouds with neural networks. 
	In section~\ref{sec:models}, we compose the blocks to form a discriminative and generative model, respectively. 
	The discriminative model finds low-dimensional representations optimized for regression or classification tasks; the generative model aims to estimate the data distribution and samples from it to obtain new shapes.
	Finally, section~\ref{sec:multistructure} presents the extension of the models to the joint analysis of multiple structures.

	
	\subsection{Computational Blocks}
	\label{sec:compblocks}
	We present four computational blocks for shape analysis on point clouds. The input are point cloud representations $\mathbf{P}= \{\mathbf{p}_1,...,\mathbf{p}_N\}$ with $\mathbf{p}_i=[x_i,y_i,z_i]$ being the coordinates of point~$i$ on the surface of an anatomical structure and $N$ the number of points.
	
	\subsubsection{Global Signature Network} \label{sec:basenetwork}
	A key building block of our method is a function  $f: \mathbf{P} \mapsto \mathbf{v}$, which maps a point cloud representation to a \emph{global shape signature} $\mathbf{v} \in \real^{F}$. Function $f$ has to take the properties of point clouds into account. A point cloud representation is \emph{unordered}, which means that any operation applied to the point cloud $\mathbf{P}$ must be invariant to permutations on the ordering of the points. \revision{This also ensures that we do not need one-to-one correspondences between point clouds.}
	In our framework, function $f$ is based on a PointNet network \citep{Qi2017}, which is an architecture specifically designed to operate on point cloud representations. The architecture of the network,  which approximates function $f$ is based on two operations:
	
	\begin{figure}
		\centering
		\includegraphics[width=\linewidth]{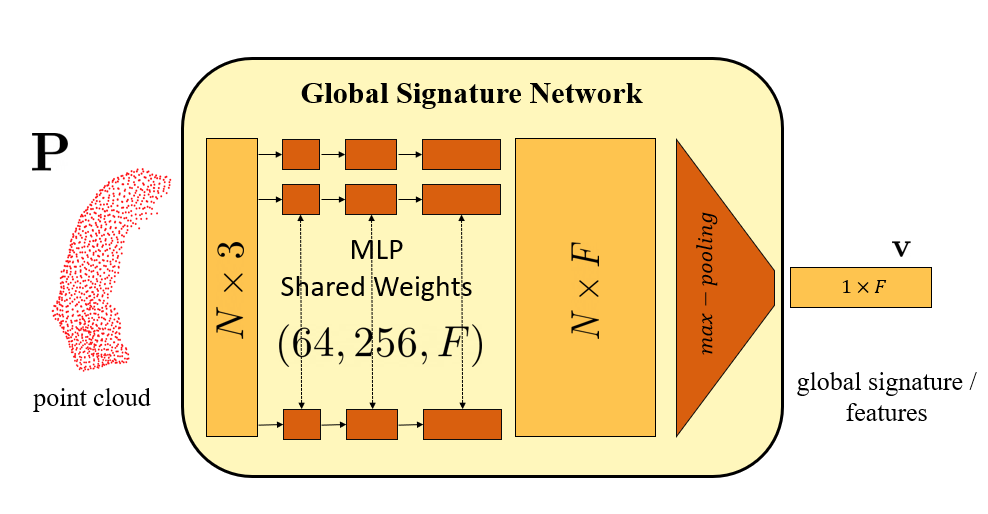}
		\caption{The global signature network is the main building block of our approach, which maps a point cloud $\mathbf{P}$ to the global signature $\mathbf{v}$. The network is composed of a multi-layer perceptron (MLP) with shared weights and a max-pooling operator. }
		\label{fig:base}
	\end{figure}
	
	\begin{itemize}
		\item A function $h : \mathbf{p}_i \mapsto \mathbf{h}_i$, which maps each individual point $\mathbf{p}_i$ to a higher dimensional representation $\mathbf{h}_i \in\real^F$. By applying $h$ to each point of the cloud $\mathbf{P}$, we obtain the matrix $\mathbf{H}\in\real^{N\times F}$, which corresponds to a high dimensional representation of point cloud $\mathbf{P}$.
		\item A symmetric operator $g : \mathbf{H} \mapsto\mathbf{v}$, which aggregates these high dimensional representations into a vector $\mathbf{v} \in\real^F$ that corresponds to a \emph{global signature} of the point cloud; in practice, function $g$ corresponds to a max-pooling operator $g(\mathbf{H}) := \left[\max_{i \in N} \mathbf{H}_{ij} \right]_{j=1,\ldots,F}$. 
	\end{itemize}
	
	\noindent
	Putting these together, function $f: \mathbf{P} \mapsto\mathbf{v}$ is approximated by:
	
	\begin{equation}
	\mathbf{v} = f(\mathbf{P}) \approx g(h(\mathbf{P})). 
	\end{equation}
	
	Fig. \ref{fig:base} illustrates the Global Signature Network (GSN) that approximates function $f$. 
	It consists of a \emph{multi-layer perceptron} (MLP) that projects the point cloud $\mathbf{P}$ ($N \times 3$) into a higher dimensional representation ($N \times F$). 
	The subsequent \emph{max-pooling} layer extracts the most relevant features from this representation to form the global signature $\mathbf{v}$ ($1 \times F$). 
	\revision{\cite{Qi2017} also evaluated to learn the pooling operator, however, the \emph{max-pooling} layer yielded higher performance so that we use it in our network.}

	\revision{GSN blocks are components for the networks  in the following sections, where the dimensionality $F$  depends on the complexity of the task (e.g., $F=256$ for the rotation network in \ref{sec:rotationnetwork} and $F=1024$ for the discriminative network  in \ref{sec:encoder}). }
	
	\subsubsection{Rotation Network} \label{sec:rotationnetwork}
\begin{figure}[t]
		\centering
		\includegraphics[width=\linewidth]{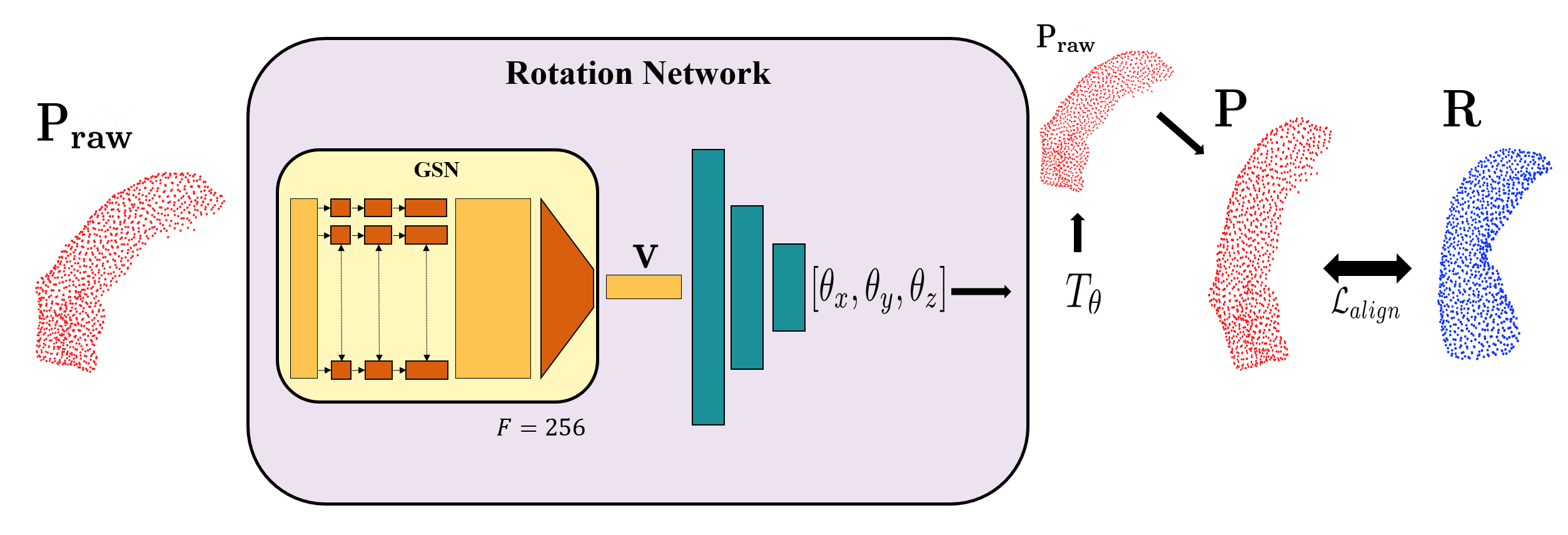}
		\caption{\revision{The rotation network transforms the input point cloud $\mathbf{P}_{raw}$ to bring it into alignment with the reference template $\mathbf{R}$. It is composed of a GSN block, which extracts the signature vector $\mathbf{v}$, and an MLP, which regresses the rotation parameters $\theta$ from the signature~$\mathbf{v}$. 
		The original point cloud is aligned to a template $\mathbf{R}$ by applying the transformation $\mathbf{T_\theta}$, parameterized by $\theta$.
		The quality of the alignment is measured by the loss function $\mathcal{L}_{align}$.} }
		\label{fig:rotation}
	\end{figure}

	According to one of its most popular definitions, shape is all the geometrical information that remains when location, scale and rotational effects are filtered out from an object \citep{kendall1989survey}.
	Hence, when our network receives as input a raw point cloud $\mathbf{P}_{raw}$, we must first ensure that its output is invariant to similarity transformations (scaling, translation, and rotations). 
	In our framework, the effects of scale and translation are eliminated by centering all shapes around their center of mass, and by normalizing the range of the coordinates of the points to lie within the range $[-1,1]$. 
To guarantee invariance to rotation, we introduce a rotation network  that learns the mapping $\mathbf{P}_{raw} \mapsto \theta$, such that $\mathbf{P} = \mathbf{T(\theta)}\mathbf{P}_{raw}$ is in spatial alignment with a reference point cloud $\mathbf{R}$. 
\revision{A similar idea is also used by spatial transformer networks~\citep{jaderberg2015spatial,he2019spectral}.}
The rotation matrix $\mathbf{T(\theta)}$ is parameterized by the rotation vector $\theta=[\theta_x, \theta_y, \theta_z]^T$.
Fig. \ref{fig:rotation} illustrates the rotation network, which builds onto the GSN to extract the global signature vector $\mathbf{v}$ that is then fed to a multilayer perceptron to regress the rotation vector $\theta$. 
	

To ensure the alignment with the reference point cloud, we measure a distance between $\mathbf{R}$ and $\mathbf{\mathbf{P}}$. 
As we are operating on unordered point clouds, we require a metric, which is permutation invariant. 
We use the 1-Wasserstein distance, also known as earth mover's distance (EMD) \citep{rubner2000earth}, in our loss function
\begin{equation} \label{eq:emd}
	\mathcal{L}_{align}(\mathbf{P},\mathbf{R}) =  EMD(\mathbf{P},\mathbf{R} ) =   \min_{\phi:  \mathbf{P} \rightarrow \mathbf{\hat{R}}} \sum_{\mathbf{p}\in\mathbf{P} } || \mathbf{p} - \phi(\mathbf{p})||_1, 
\end{equation}
where $\phi(\mathbf{p})$ is the optimal bijection that maps a point $\mathbf{p} \in \mathbf{P}$ to a point \revision{point $\mathbf{r}\in\mathbf{R}$, such that the distance between the two sets is minimal. Notice that EMD differs from other point cloud measures, based on Nearest Neighbour search (e.g. Chamfer Distance or ICP error), in the fact that it enforces a one-to-one mapping between two sets of points}.

	

	\subsubsection{Encoder Network}
	\label{sec:encoder}
	
		\begin{figure}[t]
		\centering
		\includegraphics[width=\linewidth]{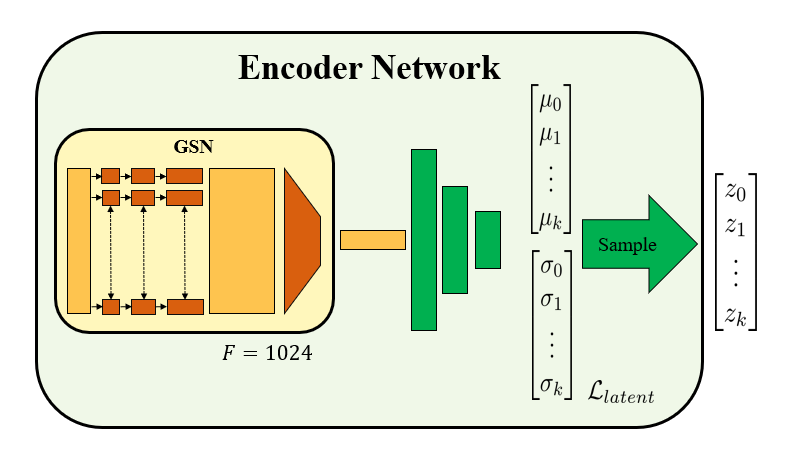}
		\caption{The encoder network finds low-dimensional shape representation $\mathbf{z}$ of a point cloud. 
		The global signature is extracted by the GSN and an MLP estimates the mean and variance of the data distribution. 
		The latent vector $\mathbf{z}$ is obtained by sampling from the distribution. The quality of the distribution estimation is assured by enforcing $\mathcal{L}_{latent}$, the latent loss of the variational autoencoder  given by the Kullback-Leibler divergence.
		}
		\label{fig:encoder}
	\end{figure}
	
	The \emph{encoder} block allows to produce representations~$\mathbf{z}$, which capture the geometrical information of the object of interest. 
	As such, this can be used in a generative, as well as, a discriminative model. 
	Here, we describe a more generic version of the encoder, as it would be used in a generative model. 
	The goal is to 
	approximate the posterior distribution $q(\mathbf{z}|\mathbf{P})$ with $\mathbf{z}\in\mathcal{Z}\subset\real^{k}$ being a vector in the latent space $\mathcal{Z}$. 
	For simplicity, we make the assumption that $q$ is a Gaussian probability density $\mathcal{N}(\mu, \Sigma)$, with mean $\mathbf{\mu}$ and covariance matrix $\Sigma=\text{diag}[\sigma_0,\sigma_1,...\sigma_k]$. 
	Fig. \ref{fig:encoder} illustrates the encoder network that approximates $q$. The network is based on the GSN to compute the global signature, which is the input of an MLP with output $(\mathbf{\mu}, \Sigma)$. A latent vector $\mathbf{z}$ is obtained by sampling from the distribution $q$.
	The encoder bears similarities to the rotation block, but instead of estimating the rotation vector, the mean and variance of the data distribution are estimated. 

The latent loss $\mathcal{L}_{latent}$ of the variational autoencoder is given by the Kullback-Leibler divergence between $\mathcal{N}(\mu,\Sigma)$ and a Gaussian prior $\mathcal{N}(0,\mathbf{I})$. Since $\Sigma$ is a diagonal matrix, the Kullback-Leibler divergence between these distributions is
	
	\begin{equation} 
	\mathcal{L}_{latent} = \sum_{i=1}^k \sigma_i + \mu_i - \log(\sigma_i) - 1.
	\label{eq:latentLoss}
	\end{equation}

	\subsubsection{Decoder Network}
		\begin{figure}[t]
		\begin{center} 
			\includegraphics[width=0.8\linewidth]{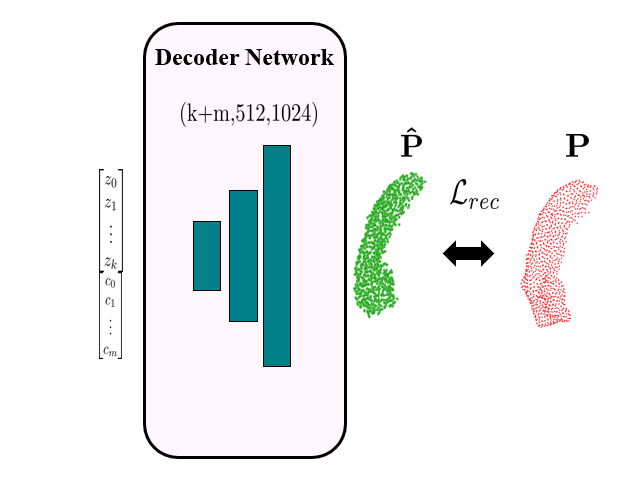}
		\end{center}
		
		\caption{Decoder network approximates the mapping  $ [\mathbf{z}, \mathbf{c}] \mapsto \mathbf{P}$. The input of the decoder are the embedding $\mathbf{z}$ and the condition vector $\mathbf{c}$. The output is the reconstructed point cloud $\hat{\mathbf{P}}$. The accuracy of the reconstruction is measured using the reconstruction loss $\mathcal{L}_{rec}$. 
		}
		\label{fig:decoder}
	\end{figure}

	The \emph{decoder} network takes as input a latent vector $\mathbf{z} \in \real^{k}$ and generates a point cloud  $\hat{\mathbf{P}}$. 
	In our framework, we consider next to the latent vector also   a condition vector $\mathbf{c} \in \real^{m}$ as input to the decoder, as illustrated in  Fig. \ref{fig:decoder}.
	The condition vector $\mathbf{c}$ is an optional set of parameters that can be used to introduce prior knowledge in the shape model. 
	For example, the condition vector can include a variable indicating the diagnostic status of the point cloud being generated. 
    The decoder therefore models the mapping $[\mathbf{z},\mathbf{c}]\mapsto \mathbf{P}$. 
	As illustrated, the decoder approximates this mapping with a fully connected MLP with 3 layers. 
	In a generative model, the decoder is assigned a reconstruction loss
	\begin{equation}
	\mathcal{L}_{rec} = EMD(\mathbf{P},\hat{\mathbf{P}}),
	\end{equation}
	which assesses the quality of the reconstructed point cloud $\hat{\mathbf{P}}$ with respect to $\mathbf{P}$.

	\subsection{Single-Structure Models}
	\label{sec:models}
	We present a discriminative and a generative model for shape analysis on single anatomical structures based on the previously introduced computational blocks. 
	
		\begin{figure}
		\centering
		\includegraphics[width=\linewidth]{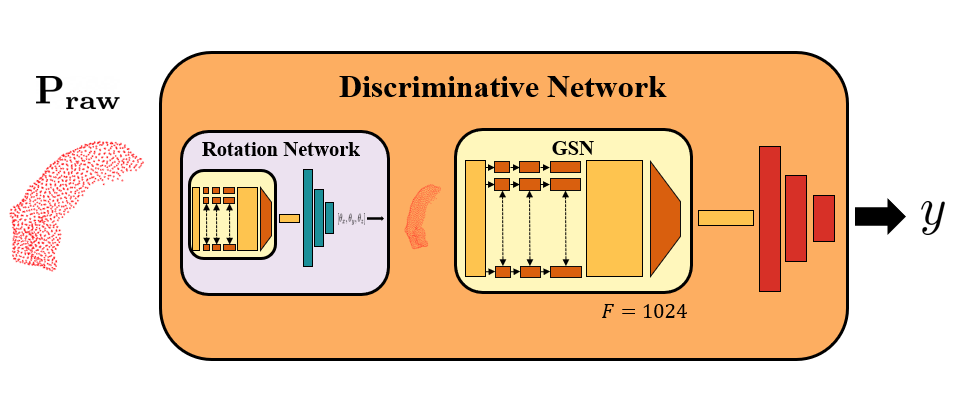}
		\caption{The Discriminative Model is based on three main components: 1) the rotation module bringing the input point cloud to a canonical space, 2) the Global Signature Network which extracts the feature vector ($F=1024$) and passes it through, 3) a fully connected block for making the final decision.}
		\label{fig:discriminative}
	\end{figure}
	
	\subsubsection{Discriminative Model}
	The goal of the  discriminative model is to learn the mapping $\mathbf{P} \mapsto y$ between a point cloud and a label $y$, which can either be a real number in the case of a regression task or a categorical variable for classification.  
	Fig.~\ref{fig:discriminative} illustrates the discriminative model that is trained in an end-to-end fashion. \revision{Note that the network consists of two GSN blocks (one inside the rotation network), which are independent.}
	
	The raw point cloud $\mathbf{P_{raw}}$ is the input to the discriminative model and first processed with rotation network to align the point cloud.
	The global signature is then extracted with the GSN and passed through multi-layer-perceptron with label $\mathbf{y}$ as output. 
	The discriminative network is trained end-to-end by combining a classification/regression loss $\mathcal{L}_{cls}$ with the alignment loss $\mathcal{L}_{align}$ 
		\begin{equation}
	\mathcal{L} = \mathcal{L}_{align} + \mathcal{L}_{cls}.
	\label{eq:loss_disc_single}
	\end{equation}
	For $\mathcal{L}_{cls}$, we could  either use a cross entropy loss for classification tasks or an $l_2$-loss in the case of regression. 
	Training the network in this manner will result in global signatures $\mathbf{v}$ of point clouds that are optimized for the particular discriminative task.

	\subsubsection{Generative Model}
	
\begin{figure*}
		\centering
		\includegraphics[width=0.7\linewidth]{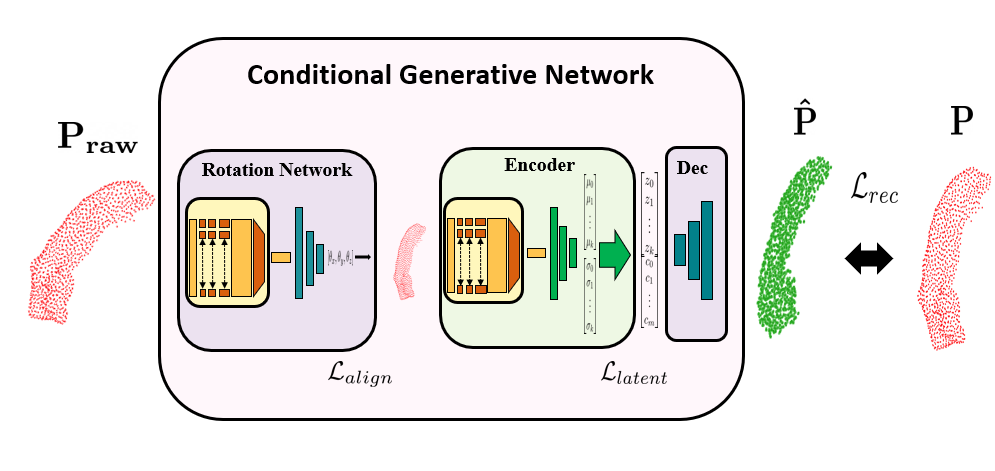}
		\caption{The Generative Model is based on a Conditional VAE. It is formed by three main components: 1) the rotation module bringing the input point cloud to a canonical space, 2) the encoder module approximating the posterior distribution $P_E(\mathbf{z}|\mathbf{P})$,  and 3) the decoder module reconstructing the point cloud by approximating the mapping $[\mathbf{z}, \mathbf{c}] \mapsto \mathbf{P}$.}
		\label{fig:generative}
	\end{figure*}

	Fig. \ref{fig:generative} illustrates the generative model, which is based on a Conditional Variational Autoencoder (CVAE). 
	The network encodes a point cloud into a $k$-dimensional latent variable $\mathbf{z}$, and then decodes this embedding to reconstruct a point cloud~$\hat{\mathbf{P}}$. 
	The generative model combines three computational blocks: 1) a rotation network, which aligns raw point clouds to a canonical space, 2) an encoder, which approximates the posterior distribution $q(\mathbf{z}|\mathbf{P})$, and 3) a decoder, which  models the mapping $[\mathbf{z},\mathbf{c}]\mapsto \mathbf{P}$ with a condition vector   $\mathbf{c}$. 
	
	The generative model is trained  by optimizing the combined loss function 
	\begin{equation}
	\mathcal{L} = w_{a} \mathcal{L}_{align} + w_{r}\mathcal{L}_{rec}   + w_{l}\mathcal{L}_{latent},
	\label{eq:loss_gen_single}
	\end{equation}
	with the alignment ($\mathcal{L}_{align}$),  reconstruction  ($\mathcal{L}_{rec}$) and latent ($\mathcal{L}_{latent}$) losses. \revision{The weights $w_a$, $w_r$ and $w_l$ balance the contribution of each one of these terms in the total loss function. In practice, the most important weight parameters to take into account is the weight of the latent loss $w_a$. A  small weight of the latent loss removes the variational component of the autoencoder, and turns the model into a regular autoencoder. As a consequence there is no way to ensure that the learned representations are smooth. However when the weight of the latent loss is too high, the variational autoencoder looses its ability to capture high frequency details of the shape. In our experiments the values of these weights are $w_a=1$,$w_r=1$, $w_l$=10}.
	
	The resulting generative model finds a low-dimensional shape representation $\mathbf{z}$ given an input point cloud, which could be used as features for training a classifier, as we will describe in section \ref{section:exp_AD}. At the same time, the model allows to generate point clouds $\mathbf{\hat{P}}$ from the learned embedding space by sampling $\mathbf{z}$ from a multivariate Gaussian and setting a condition vector $\mathbf{c}$.

	\subsection{Multi-Structure Models} \label{sec:multistructure}
	So far, we have focused on shape analysis for single structures. 
	An advantage of our framework is that it can be easily extended to the joint analysis of multiple structures. 
	This is important as we are often interested in a holistic morphological modeling of multiple anatomical structures, which are not independent from each other. 
	To ease notation, we use $\mathbf{P^0}, \ldots, \mathbf{P^K}$ to refer to multiple raw point clouds without explicitly stating `raw'. 
	
	\subsubsection{Multi-structure Discriminative Model}
\begin{figure}[t]
		\centering
		\includegraphics[width=\linewidth]{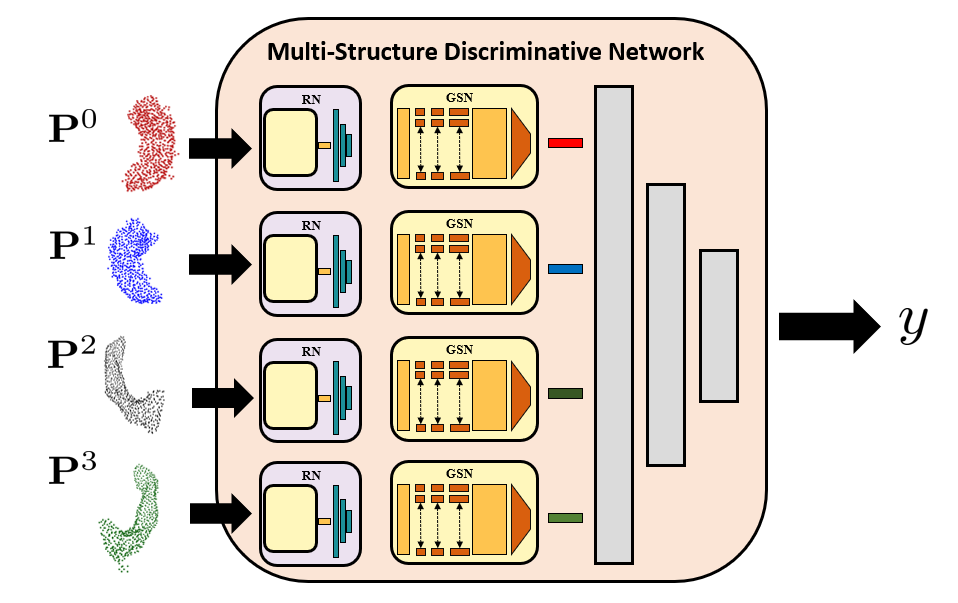} 
		\caption{In the Multi-Structure Discriminative Network, each structure is processed by using separate paths to create independent global signatures, illustrated in different colors. 
		The individual signatures are then concatenated and passed through an MLP for predicting the label $y$.}
		\label{fig:msp_discriminative}
	\end{figure}

	Fig.~\ref{fig:msp_discriminative} illustrates the multi-structure extension of the discriminative model. 
	Each shape is first passed through an individual rotation network and GSN to extract a global signature per shape. 
	These global signatures are then concatenated and fed into an MLP to make the prediction, which takes into account morphological features of all structures. 
	As an alternative to this architecture, we could directly concatenate all point clouds and pass them through a single rotation network and GSN. 
	However, we have demonstrated in our earlier work \citep{gutierrez2018deep} that keeping separate branches for each individual structure yields to a higher performance than directly concatenating point clouds. 

	\subsubsection{Multi-structure Generative Model} 
\begin{figure*}[t]
		\centering
		\includegraphics[width=0.9\linewidth]{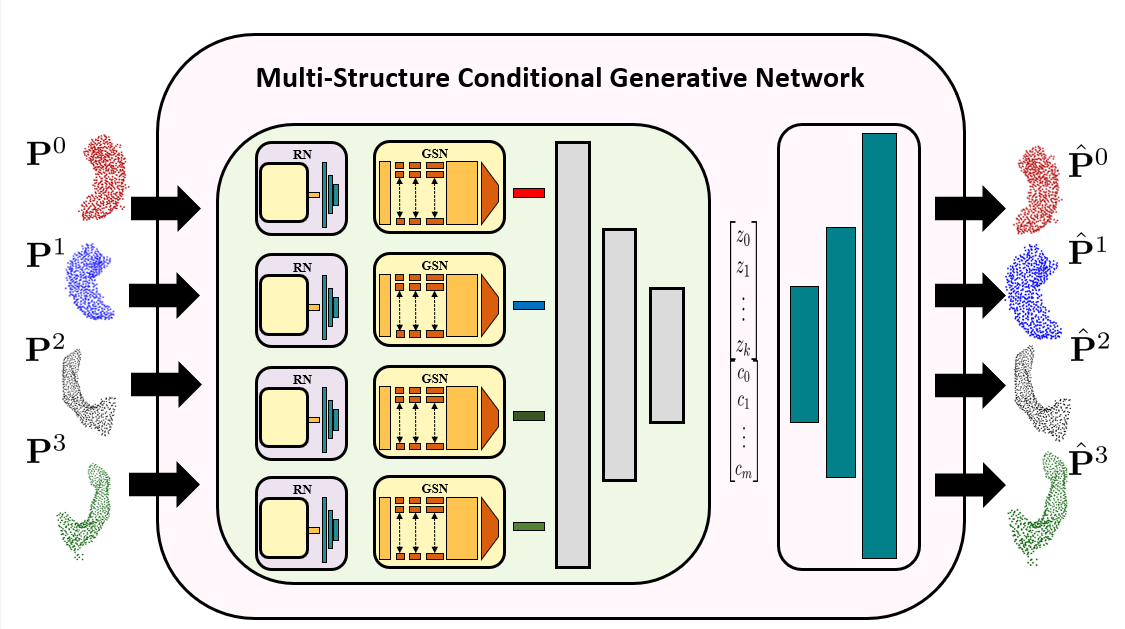}
		\caption{In the Multi-Structure Generative Network, each structure is processed using separate paths to create independent global signatures. 
		The signatures are then concatenated and fed into an MLP to estimate the posterior probability $q(\mathbf{z}|\mathbf{P^0,P^1,\dots,P^K})$.
		Notably, the embedding jointly encodes all structures. 
		For generating structures $\mathbf{\hat{P}^0,\hat{P}^1,\dots,\hat{P}^K}$, a latent vector $\mathbf{z}$ is sampled from the distribution, combined with a condition vector, and passed through the MLP decoder.  }
		\label{fig:msp_generative}
	\end{figure*}

	Fig.~\ref{fig:msp_generative} illustrates the multi-structure generative model, which concatenates independent GSN paths and passes them through an MLP in order to estimate the posterior distribution $q(\mathbf{z}|\mathbf{P^0,P^1,\dots,P^K})$. 
	The latent vector $\mathbf{z}$ is sampled from the distribution and passed through a decoder shared by all the structures.   $\mathcal{L}_{latent}$ remains identical to equation  \ref{eq:latentLoss}, while $\mathcal{L}_{rec}$ and $\mathcal{L}_{align}$ are averaged over the different structures
		\begin{equation}\label{eq:emd_avg}
	    \mathcal{L}_{rec} = \frac{1}{K+1}\sum_{i=0}^{K} EMD(\mathbf{P^i,\hat{P^i}}),
	\end{equation}
	where the extension of $\mathcal{L}_{align}$ is analogously.

	\section{Experiments}
	We evaluate the discriminative and generative models in several supervised and unsupervised tasks, where we describe the data and pre-processing in section~\ref{section:data}. In section \ref{section:exp_AD}, we analyze the performance of the  discriminative model in regression and classification tasks. In sections \ref{sec:experiments_gen_single} and \ref{sec:experiments_gen_mult}, we evaluate the generative model for  single and multi-structure tasks.

\subsection{Data and pre-processing}
\label{section:data}
We want to explore the capabilities of our model in different scenarios. Hence, in our experiments, we work with point clouds from two datasets from different domains. As explained before, all  point clouds   are centered and \revision{divided by the Euclidean norm of the furthest point (with respect to the origin of coordinates), so that their values lie inside a sphere with radius 1}, in order to remove the effects of translation and scale.

\subsubsection{ADNI}
We work with data from the Alzheimer's Disease Neuroimaging Initiative (ADNI) (adni.loni.usc.edu)~\citep{Jack2008}, which contains magnetic resonance imaging (MRI) scans from healthy controls (HC) and patients with mild cognitive impairment (MCI) and  Alzheimer's disease (AD). 
Segmentations of the structures of interest are obtained using the automatic segmentation tools included in  FreeSurfer \citep{Fischl2012}. From these segmentations meshes are extracted and point clouds are obtained by uniformly sampling points for each structure. 
ADNI is a longitudinal study, i.e., we have multiple follow-up scans per participant. In total, we work with 7,974 images (2,423 HC, 978 AD, and 4,625 MCI). 
\revision{As reference shape for the rotation network, we randomly selected an image from the IXI dataset (IXI013), where we noted similar results for different reference shapes.}

\subsubsection{3D-MNIST}
As second dataset, we use a 3D point cloud version of the MNIST database\footnote{https://www.kaggle.com/daavoo/3d-mnist}. 
It contains point clouds of handwritten digits from 0 to 9. 
The advantage of this computer vision dataset is that the classes are visually very different, so that it is easier to perceive the model's performance. In contrast, changes on medical data are typically much smaller and therefore not as easy to visualize. 
The dataset contains 5,000 3D point clouds, where we uniformly sampled 1,024 points for each point cloud.

\subsection{Discriminative Predictions}\label{section:exp_AD}
	In our first experiment, we perform regression and classification tasks  on shapes from ADNI. 
	In the classification task, \revision{we perform two experiments: in the first one we classify between healthy controls and patients diagnosed with Alzheimer's Disease (AD) and in the second one we perform classification between healthy controls and patients with Mild Cognitivie Impairment (MCI).} 
	In the regression task, we estimate the age. 
	For these experiments, we use shapes from 4 brain structures and 512 points per structure: the left and right hippocampi and the left and right lateral ventricles. We split the data in training, validation and test sets (70\%, 15\%, 15\%) on a per subject basis in order to guarantee that scans from the same subject do not appear in different sets.
	
		
		
		
		
		

		\begin{table}
	    \centering
		\begin{tabular}{l r r r}
			\toprule
			 \ & \ Precision \ & \ Recall \ & \ F1-score  \\
			\midrule
			\multicolumn{4}{l}{\textbf{HC-AD Classification}} \\
			BrainPrint & $0.80 $ & $0.79$ & $0.78$  \\
			Generative Model & $0.78$ & $0.79$ & $0.78$   \\
			Discriminative Model & $\mathbf{0.83}$ & $\mathbf{0.84}$ & $\mathbf{0.82}$   \\
			\midrule
		
		\multicolumn{4}{l}{\revision{\textbf{HC-MCI Classification}}} \\
		\revision{BrainPrint} & $0.60 $ & $0.69$ & $0.62$  \\
			\revision{Generative Model} & $0.59$ & $\mathbf{0.72}$ & $0.63$   \\
			\revision{Discriminative Model} & $\mathbf{0.66}$ & $0.68$ & $\mathbf{0.67}$   \\
			\bottomrule 
		\end{tabular}
		
		\label{tab:results_wrt_cls}
		\caption{\revision{Precision, Recall and F1-score for AD and MCI classification with BrainPrint, the generative model, and the discriminative model. 
		For BrainPrint and the generative model, an MLP classifier is trained on the shape representations.} 
		}
	\label{table:classification}
	\end{table}

	


	\begin{figure}[t]
		\centering
		\includegraphics[width=\linewidth]{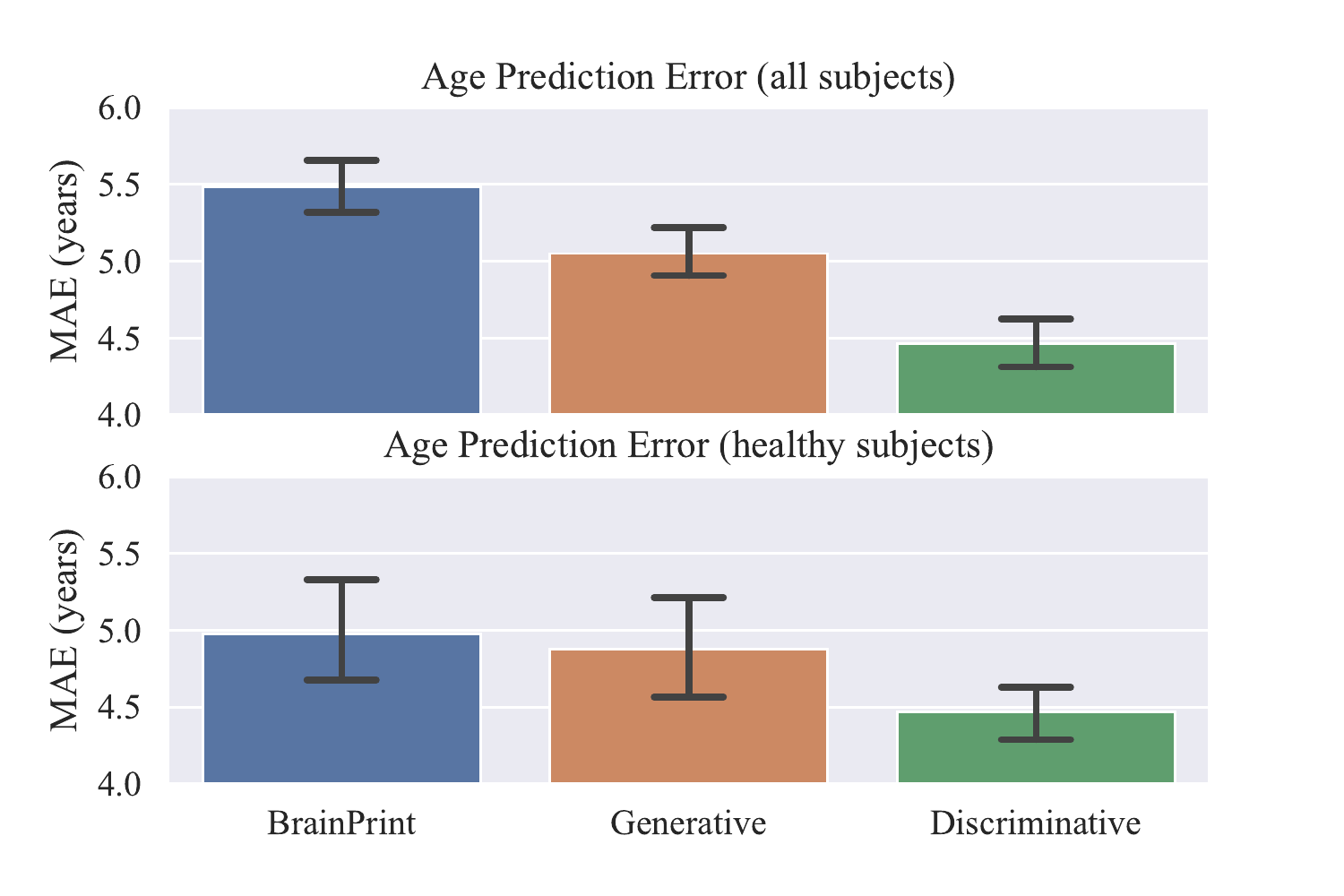}
		\caption{Mean absolute error (MAE), and its standard deviation, in years for the age prediction experiments with all ADNI subjects (top) and healthy controls (bottom). We compare BrainPrint, the generative model, and the discriminative model. 
		For BrainPrint and the generative model, an MLP  is used for obtaining a prediction from the shape representations. 
		}
		\label{fig:regression}
	\end{figure}
	We evaluate three different methods to perform the discriminative tasks: 1) a multilayer perceptron  trained on BrainPrint features \citep{Wachinger2015}, which are shape descriptors  that have presented high performance on Alzheimer's disease classification and age estimation~\citep{wachinger2016domain}, 2) a multilayer perceptron trained on shape features obtained using the multi-structure autoencoder (i.e. $\mathbf{z}$ in Fig.~\ref{fig:msp_generative}), and 3) the end-to-end multi-structure discriminative network, as illustrated in Fig.~\ref{fig:msp_discriminative}.

		\begin{figure*}[!t]
		\centering
		\begin{subfigure}[t]{0.44\textwidth}
			\includegraphics[width=\textwidth]{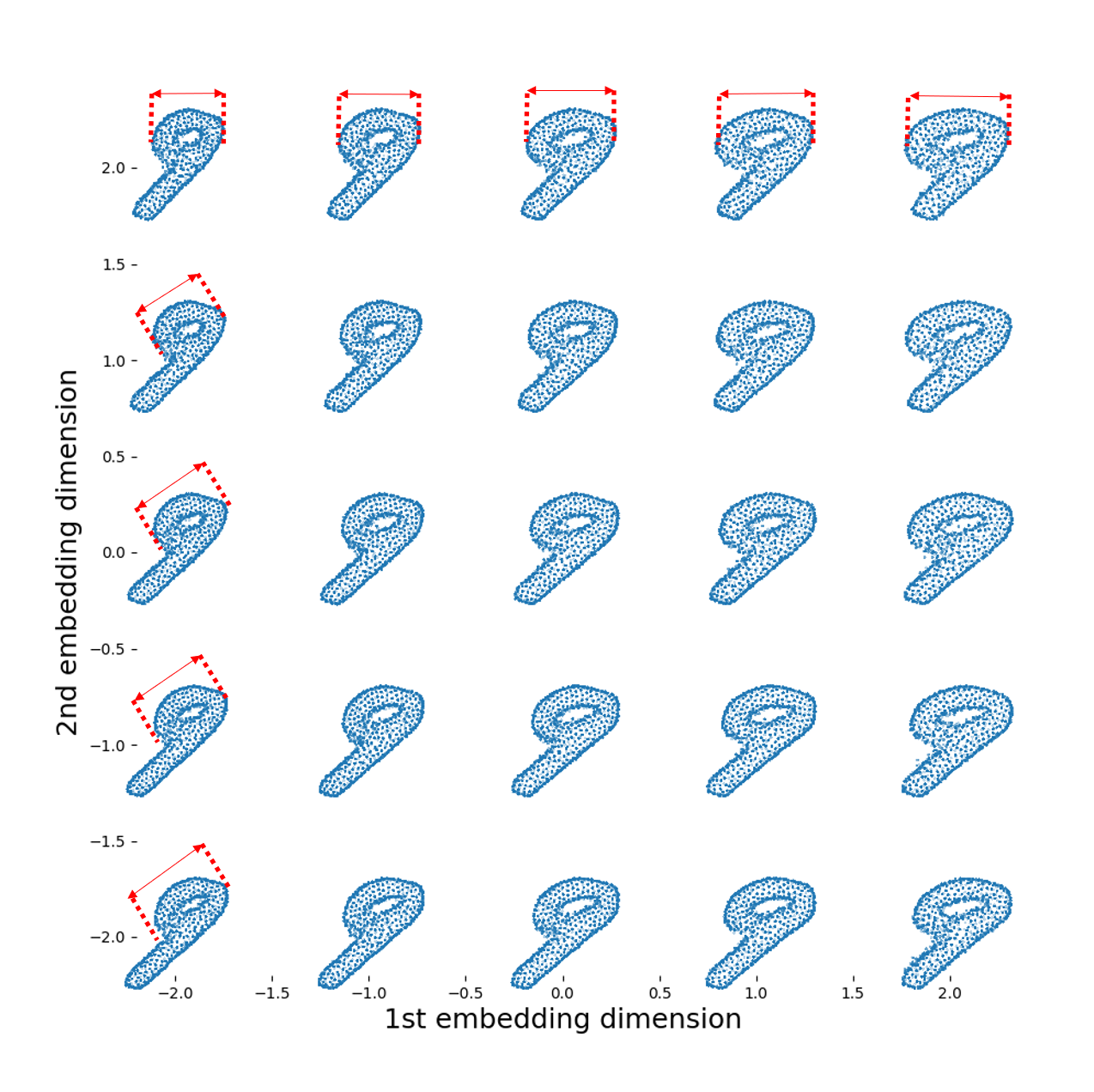}
			\caption{\revision{Conditional vector set to generate the number 9.}}
		\end{subfigure}
		~
		\begin{subfigure}[t]{0.44\textwidth}
			\includegraphics[width=\textwidth]{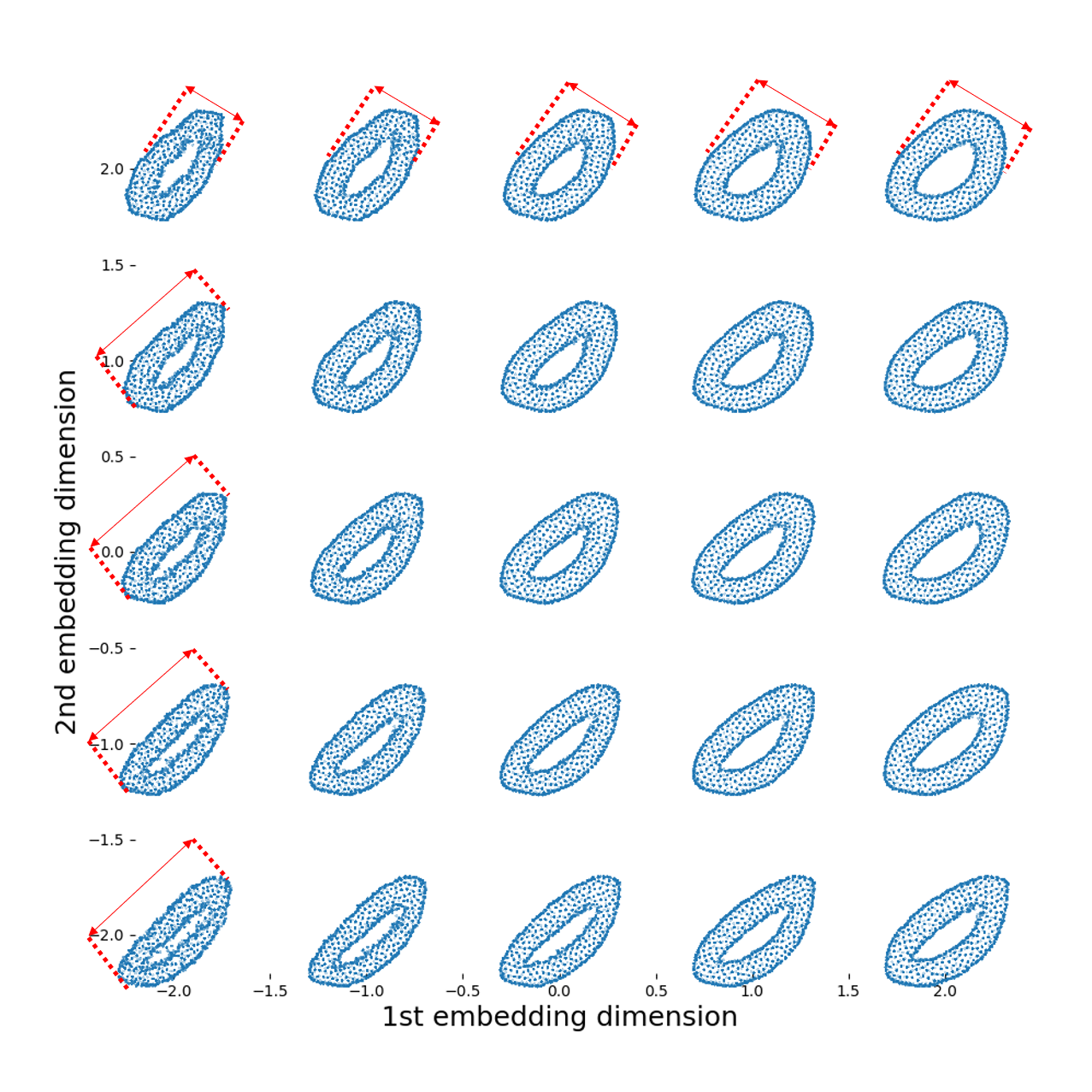}
			\caption{\revision{Conditional vector set to generate the number 0.}}
		\end{subfigure}
		~
		\begin{subfigure}[t]{0.44\textwidth}
			\includegraphics[width=\textwidth]{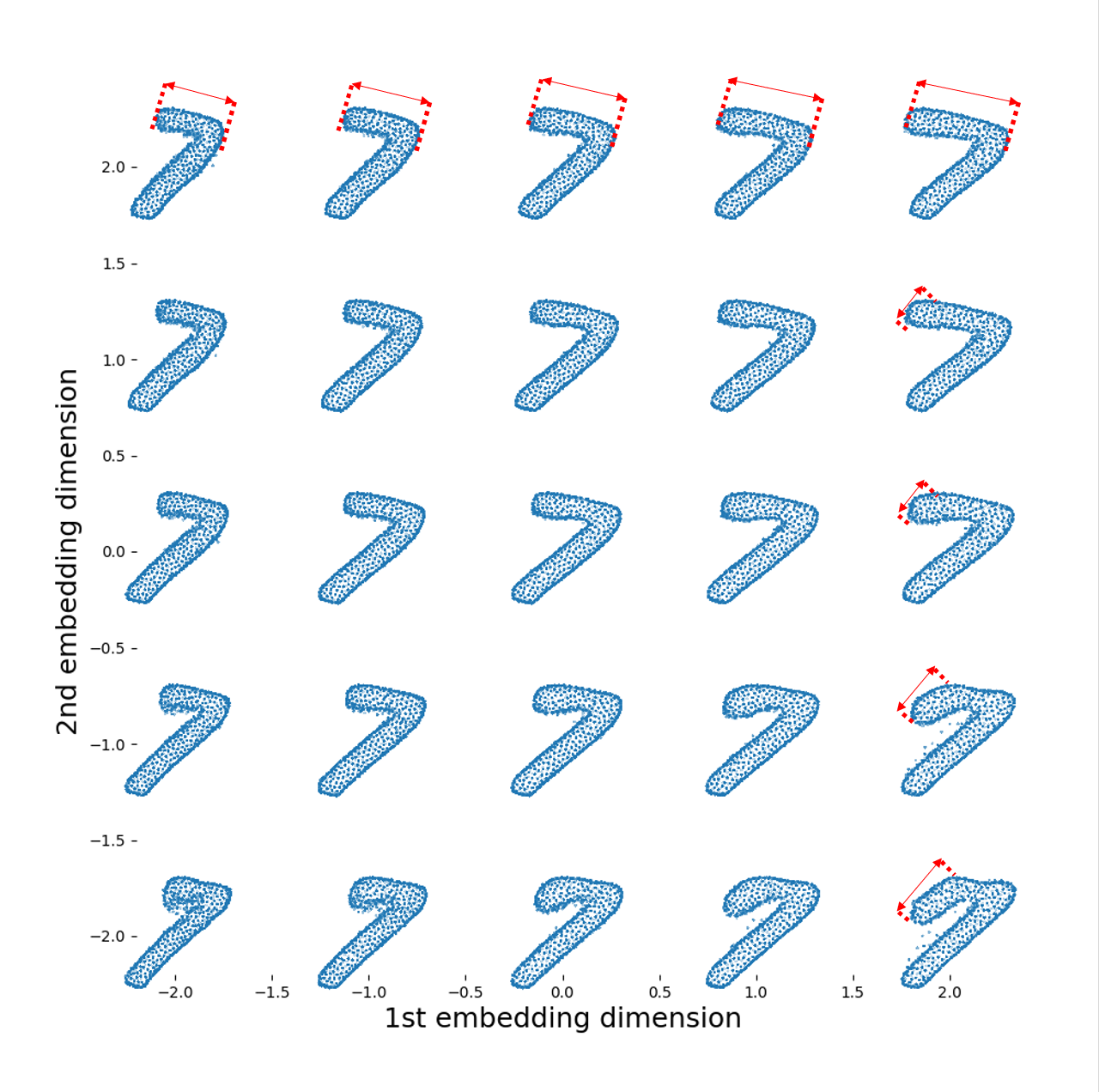}
			\caption{\revision{Conditional vector set to generate the number 7.}}
		\end{subfigure}
		~
		\begin{subfigure}[t]{0.44\textwidth}
			\includegraphics[width=\textwidth]{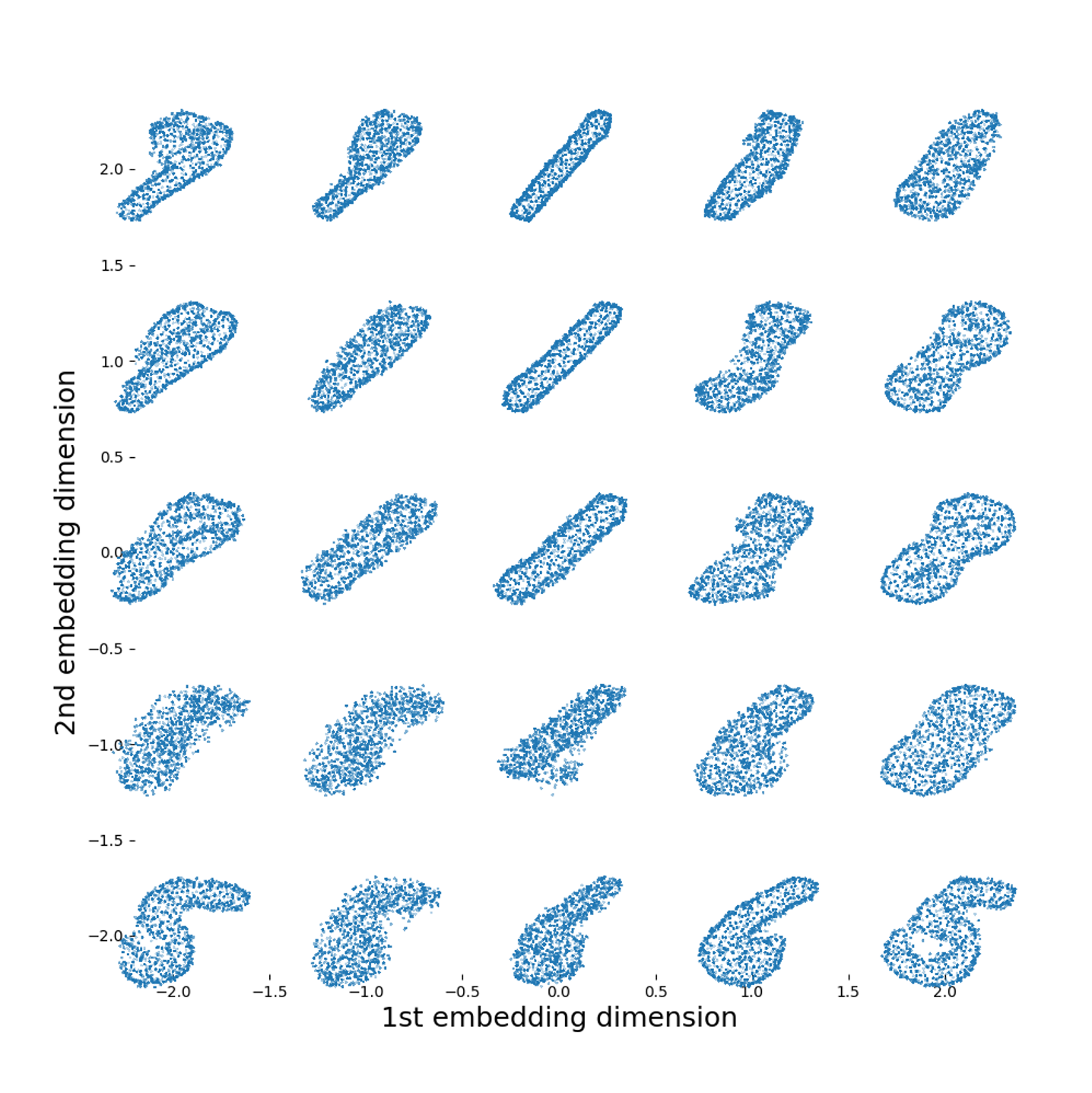}
			\caption{\revision{Conditional vector set to all zero (no condition).}}
		\end{subfigure}
		\caption{Point clouds sampled from the 2D embedding space generated by training our model using the 3D MNIST dataset. On the bottom right we show 3D point clouds generated by setting the conditional vector $\mathbf{c}$ to zero. For the other figures, $\mathbf{c}$ is set to generate point clouds of  the digits $9,0$ and $7$. \revision{The arrows show the areas where we have observed the major changes.}}
		\label{fig:mnist_grid}
	\end{figure*}

	Table \ref{table:classification} summarizes the results of the \textbf{Alzheimer's Disease Classification} experiments. \revision{For both tasks, BrainPrint and features obtained from the multi-structure autoencoder have a comparable performance, which shows that the generative model is able to capture a meaningful representation. The discriminative model has the highest precision and F1-score for both tasks and higher recall on AD classification. This illustrates the potential of learning feature representations, which are optimal for a particular discriminative task. We also observe a drop in the performance of the models for  HC-MCI classification. This is due to the high variability within the class, since the detection of MCI is more symptomatic and it is sub-divided in different stages.} 

For the regression task, we evaluate \textbf{brain age prediction}~\citep{franke2010estimating,becker2018gaussian}, i.e., the prediction of a person's age from a brain MRI scan. 
The prediction of the brain age is of interest as it was demonstrated that brain age relates to cognitive aging and that the difference to the chronological age is associated to neurodegenerative diseases. 
We perform two experiments on the ADNI dataset.
In the first one, we include all subjects from the dataset and in the second one, we only select healthy controls for the analysis. 
Fig.~\ref{fig:regression} summarizes the results of this experiment with plots of the mean absolute error (MAE). 
The evaluations are done again on the same brain structures used for the classification task. 
In the experiment on all subjects, there is a significantly lower prediction error for the discriminative model than for the generative model ($p < 0.001$) and \revision{BrainPrint} ($p < 0.001$). Further, the difference between the generative model and BrainPrint is significant ($p < 0.001$). 
For the experiment on healthy subjects, there is no significant difference between the generative model and BrainPrint,  but the improvement of the discriminative model is again significant. 

Overall, the results for the regression task are similar to those from the classification task with the discriminative model showing the best performance. 
When using the features from the generative model, we obtain a lower error than with BrainPrint features, while it is reversed for classification. 

		\begin{figure*} [t]
		\centering
		\begin{subfigure}{0.45\linewidth}
			\includegraphics[width=\linewidth]{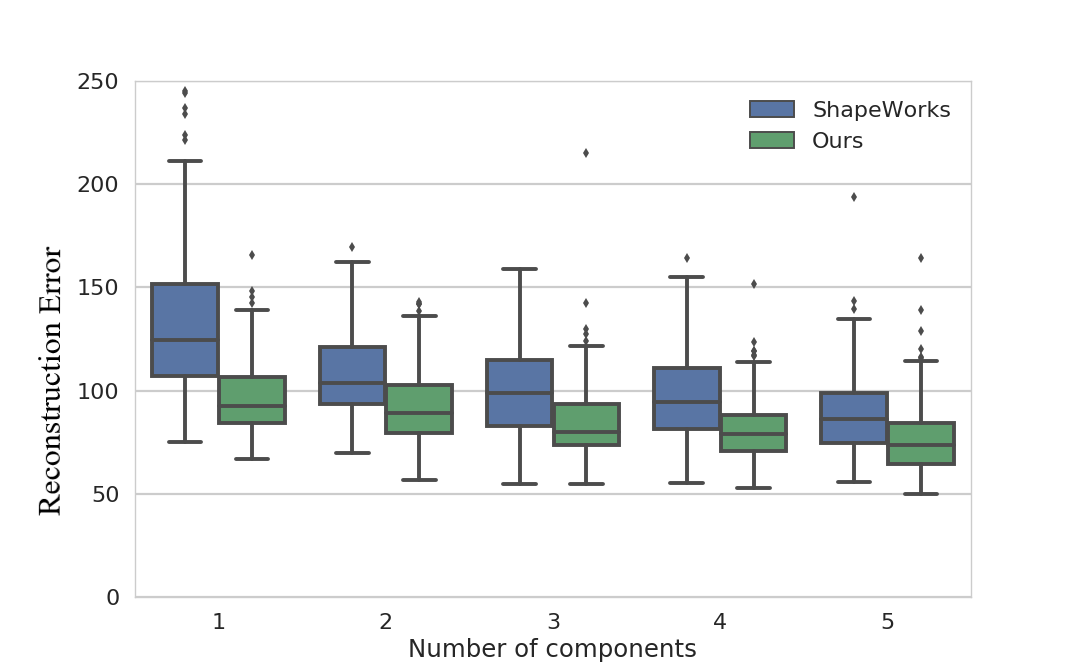}
		\end{subfigure}
		\begin{subfigure}{0.52\linewidth}
			\includegraphics[width=\linewidth]{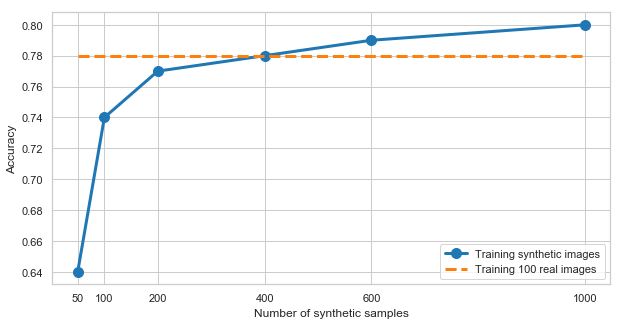}
		\end{subfigure}

		\caption{Left: reconstruction error of shapes generated using either ShapeWorks or our CVAE framework with respect to the input point cloud. Right: HC vs AD Classification accuracy for a discriminative model trained either using point clouds obtained from real segmentations  or synthetic hippocampus point clouds generated by the generative model. }
		\label{fig:reconstruction_error_accuracy}
	\end{figure*}
	

	\subsection{Single-Structure Generative Model}\label{sec:experiments_gen_single}
	In this section, we will asses the single-structure  generative model in a number of applications.
	
	\subsubsection{Conditional Shape Model of 3D-MNIST}
	As a first experiment, we train a generative shape model on 3D-MNIST and successively sample point clouds from the low-dimensional embedding. We trained two separate generative models. For the first one, we set the the dimension of the embedding $\mathbf{z}$ to $k=2$, and we use a 10-dimensional one hot encoding of the class of each digit as the condition vector $\mathbf{c}$. The second model is trained under the same settings but with the condition vector $\mathbf{c}$  set to all zeros. This means that both models are essentially identical, with the important difference that the first one is equipped with a condition vector, which allows us to give information to the network about the digit to be encoded and reconstructed. In Fig. \ref{fig:mnist_grid},  we present artificial point clouds generated by these two models. At the bottom right of Fig. \ref{fig:mnist_grid}, we  show point clouds generated without the use of the condition vector $\mathbf{c}$. Although the model is able to generate some realistically looking digits (like the 1s in the center column), the reconstructed point clouds are generally not as sharp as those generated by the conditional model. In contrast, by setting the condition vector to generate a specific digit, we are able to obtain sharp point clouds, while at the same time capturing complex non-linear deformations for each digit. The digits in Fig. \ref{fig:mnist_grid} present a very similar orientation (tilted to the right and aligned with respect to the $x,y$ plane). This is the result of aligning the point clouds to a reference template using the rotation network. An important observation is that all digits are sampled from the same shape space $\mathcal{Z}$, and only the condition vector $\mathbf{c}$ changes. This means that the encoding $\mathbf{z}$ is able to encode common shape characteristics between all digits. For example, the 1st embedding dimension in Fig. \ref{fig:mnist_grid} captures the width of the digits. It is also worth mentioning that for many typical statistical shape models, training a shape model consisting of 5,000 point clouds would be impractical due to memory limitations and to the computationally expensive task of finding corresponding points between all these shapes. 
			
		\begin{figure*}[t]
		\centering
		\includegraphics[width=0.8\linewidth]{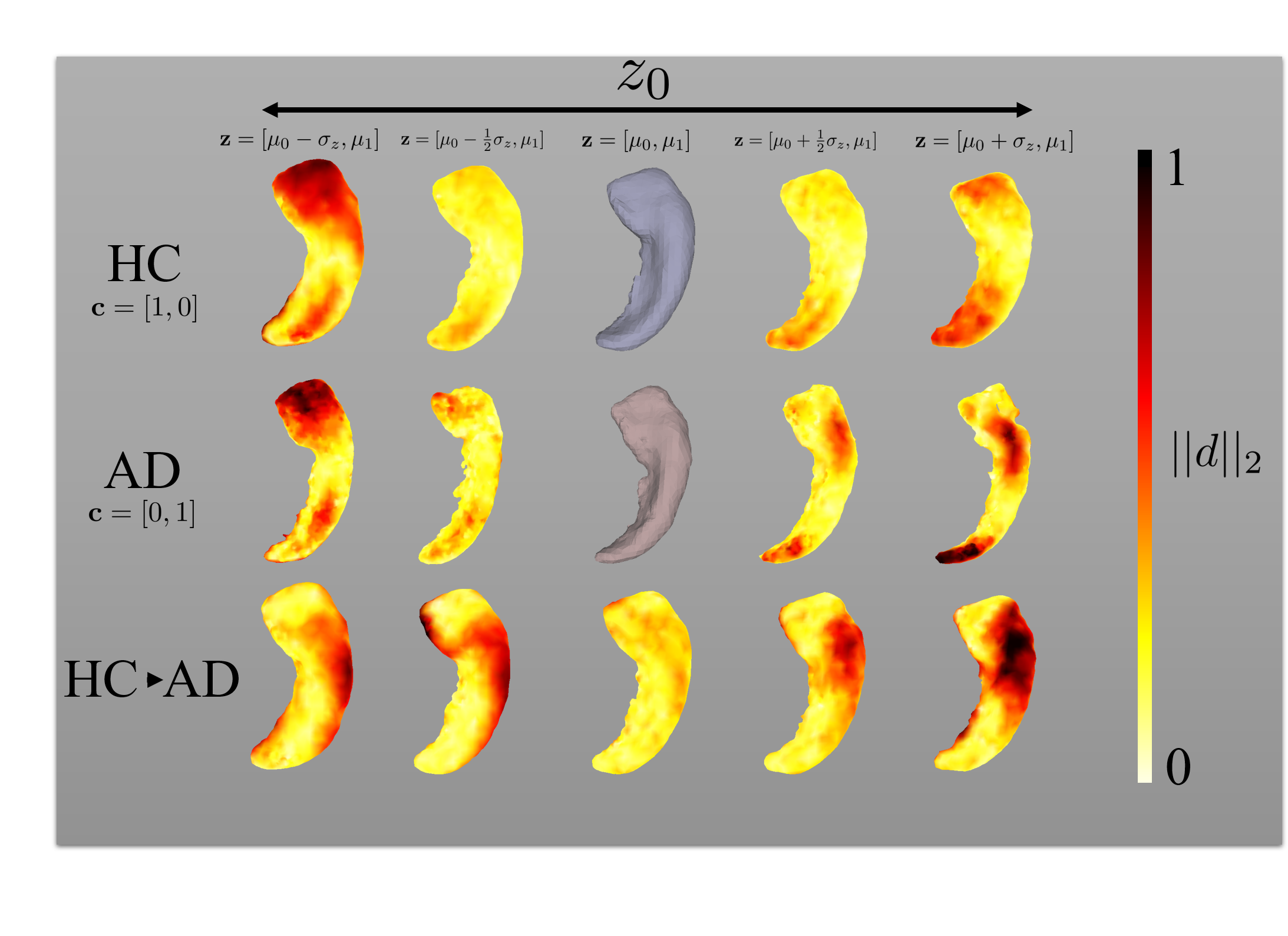}
		\caption{Hippocampus surfaces generated using point clouds sampled from our model trained on segmented images obtained from the ADNI database. The top row corresponds to point clouds generated by setting the condition vector to generate HC shapes, and the middle row corresponds to AD. Point clouds are generated by moving along the first embedding dimension. For the top two rows, the color coding shows the deformation (measured as the absolute distance between corresponding points) between the mean and the generated point cloud. In the bottom row, we show the deformations between HC and AD shapes generated using the same shape embedding $\mathbf{z}$ and only the conditional vector changes (i.e. the difference between the two top rows).}
		\label{fig:hippocampus_modes}
	\end{figure*}

	\subsubsection{Conditional Shape Model of the Hippocampus}
	In our second experiment, we build a shape model of the left hippocampus. Our goal is to assess shape differences  between healthy controls and subjects diagnosed with Alzheimer's disease. Several previous studies have established strong morphological changes in the hippocampus associated to the progression of dementia \citep{frisoni2008mapping,gerardin2009multidimensional}.   
	For comparison, we build a statistical shape model of the hippocampus using the ShapeWorks framework \citep{cates2008particle}. ShapeWorks is a statistical shape model tool, which achieved the best performance in several shape analysis tasks in a recent comparison  \citep{goparaju2018evaluation}. To perform a fair comparison with ShapeWorks, which is limited in the number of samples to be analyzed due to memory constraints, the number of samples for performing this experiment was reduced to 200 randomly selected samples (100 HC / 100 AD). We split the images into a training and testing set (100 / 100 split) and we build a statistical shape model of the left hippocampus using the training set (50 HC and 50 AD).  Segmentations are pre-processed using the grooming operations included in ShapeWorks to obtain smooth hippocampi surfaces, and models of 1024 points are trained.  As a condition vector $\mathbf{c}$, we use a one hot encoding of the diagnosis of the patient ($[0,1]$ for HC, $[1,0]$ for AD). It is also worth mentioning that training the ShapeWorks model for 100 images took 5h, compared to the 2h training time for our model. 
	
	\textbf{Reconstruction error:} 
	We first evaluate the ability of our model to obtain an accurate and compact representation of the hippocampus shape. To this end, we measure the reconstruction error between the reconstructed shapes $\hat{\mathbf{P}}$ and the input shapes $\mathbf{P}$ by evaluating $EMD(\mathbf{P},\mathbf{\hat{P}})$. We train 5 different models with embedding dimensions ranging from $k=1$ to $k=5$. As a comparison, we quantify the reconstruction error of synthetic hippocampus shapes generated by ShapeWorks. The lower reconstruction errors of our method in Fig.~\ref{fig:reconstruction_error_accuracy} indicate that it captures the complex deformations of the hippocampi and therefore allows for a compact shape representation with few modes. 
	
	\textbf{Effect of conditioning the shape model using a diagnostic label:}
	One of the main contributions that separates our framework from previous approaches for shape analysis is the introduction of the  conditional vector~$\mathbf{c}$. We have observed in our experiment on the MNIST dataset that our method is able to generate realistic shapes of digits given different condition vectors $\mathbf{c}$. 
	To evaluate the effects of the condition vector in the model of the hippocampus shapes, we use the model trained on the previous experiment (for embedding dimension $k=2$) and generate a set of synthetic  point clouds by sampling values of $\mathbf{z}$ and assigning either $\mathbf{c}=[1,0]$ or $\mathbf{c}=[0,1]$ to generate synthetic hippocampus shapes corresponding to morphological characteristics associated to either HC or AD. In Fig.~\ref{fig:hippocampus_modes} we can observe some of the synthetic shapes generated by our model, corresponding to the mean shape (center) and shapes generated by moving across the first embedding dimension $z_0$. Notice that shapes in the same column correspond were generated using the same embedding $\mathbf{z}$, with different condition vector $\mathbf{c}$. Fig.~\ref{fig:hippocampus_modes} shows that by moving across $z_0$, our model captures shape differences that are common between the HC and AD cases. For example, we observe that the left most example for both cases has a large deformation on the top part of the hippocampus. On the bottom row, we show differences between the point clouds of the top two rows, which correspond to the shape variations that our model associates to the presence of AD. These shape variations correspond to large deformations in the lateral part of the hippocampus body, roughly around the CA1 subfield. These observations are in line with previous findings on shape differences of the left hippocampus associated to AD diagnosis \citep{frisoni2008mapping,gerardin2009multidimensional}.
	
	\textbf{Synthesizing training data:}
	A critical question to answer is whether our synthetically generated point clouds capture shape differences that are specific to  AD. 
	To this end, we train our discriminative model on synethetically generated point clouds. 
	In details, our model generates hippocampi for HC and AD subjects by setting the condition accordingly and then use them for training an HC-AD classifier. 
	We vary the size of the synethetically generated training set and use the following sample sizes: 50, 100, 200, 400, 600, and 1000.
	For each dataset, a separate classifier is trained. 
	As baseline, we use a disriminative model that is directly trained on the real 100 hippocampi, which have been used for training the generative model. 
	
	Fig.~\ref{fig:reconstruction_error_accuracy} shows the classification accuracy for this experiment. 
	The results demonstrate that the generated samples are realistic enough to train a classifier relying solely on the synthetic images. 
	Interestingly, the generator allows us to sample an arbitrary number of samples, giving us the possibility to boost the accuracy of the  classifier by artificially increasing  the size of the dataset.
	This is insofar surprising, as the total amount of real data used for training is identical in both scenarios, making this a fair comparison.  


	\begin{figure}[t]
		\centering
		\includegraphics[width=\linewidth]{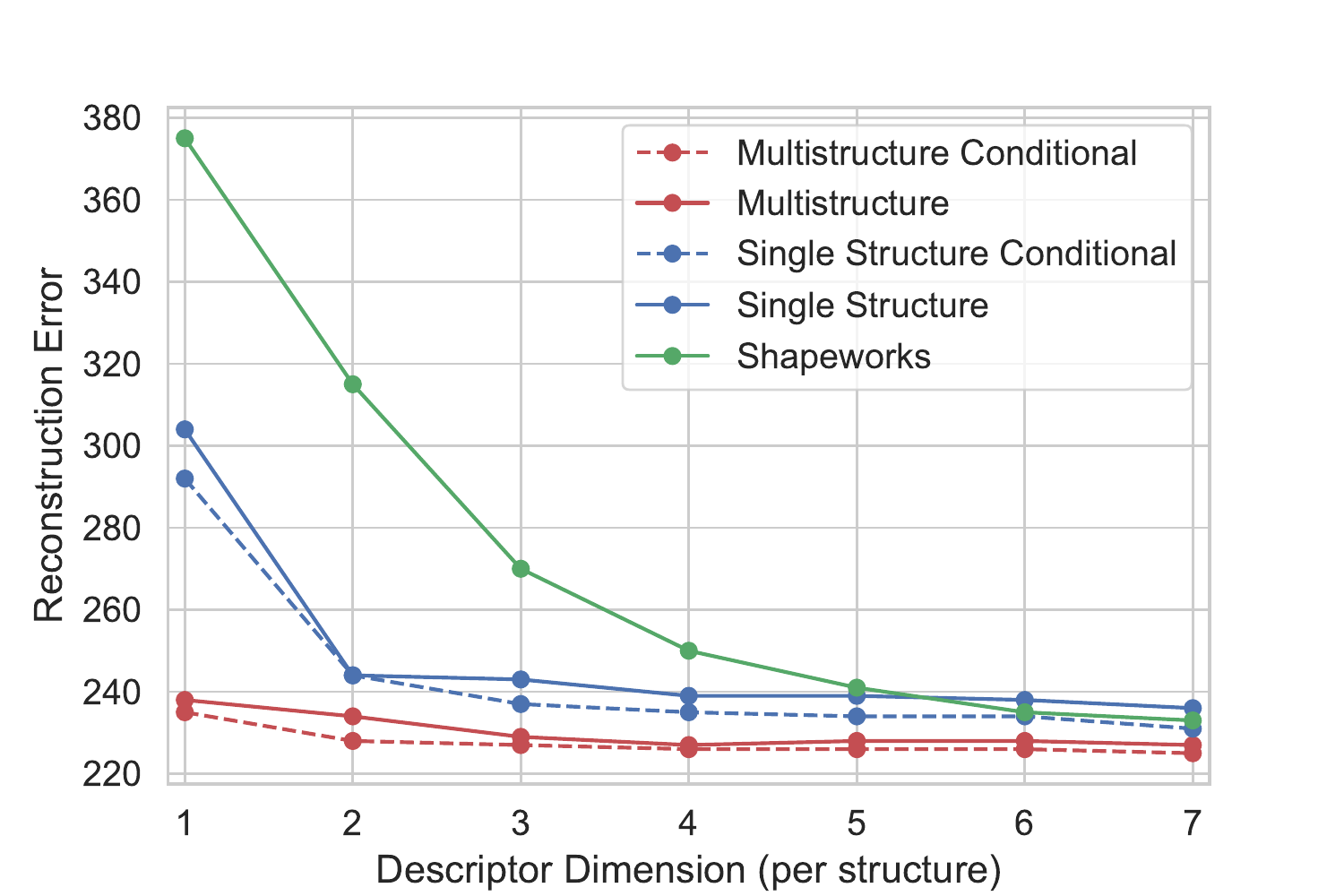}
		\caption{\revision{In order to asses the effect of conditioning and the multi-structure analysis, we compute the reconstruction error along the dimension of the feature vector per structure for different scenarios: 1) using Shapeworks for generating the individual structures (green curve) ;  2) Single Structure: Training our generative model for each structure without adding any condition (continuous blue); 3) Single Structure Conditional: training our generative model for each structure conditioning on the disease (dashed blue);  4) Multi-Structure: Training our multi-structure generative without adding any condition (continuous red); 5) Multi-structure Conditional: Training our multi-structure generative model conditioning on the disease (dashed red). Notice that for the multi-structure cases 4) and 5),  the total dimension of the descriptor is four times the dimension on the x-axis, since for a fair comparison, we analyze it per structure.} }
		\label{fig:reconstruction_error_curve}
	\end{figure}
	
		\begin{figure}[t]
		\centering
		\includegraphics[width=\linewidth]{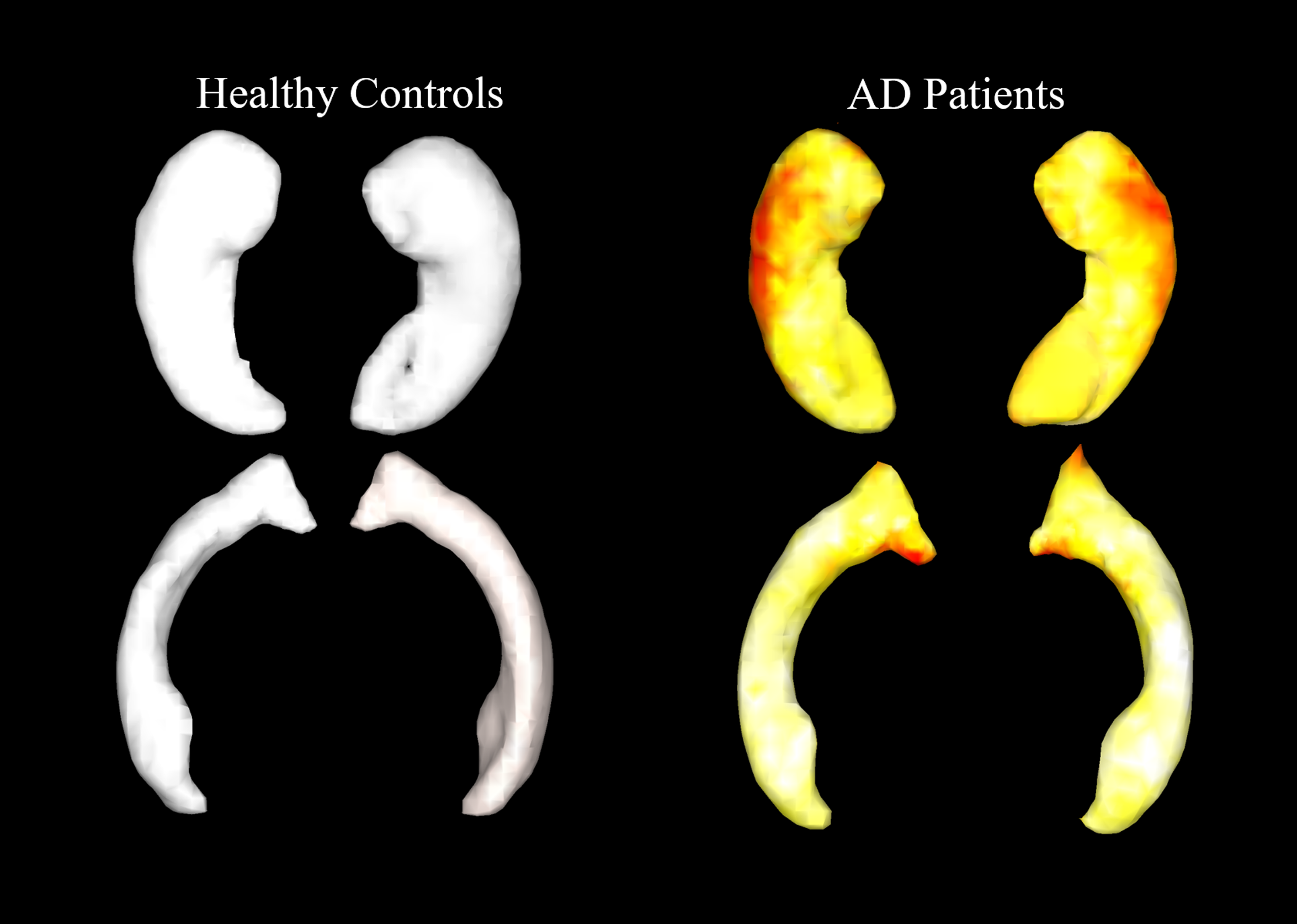}
		\caption{Deformation with respect to the healthy control (HC) mean shapes when only the conditional vector changes from HC to AD. Top row corresponds to left and right hippocampi and the bottom to left and right lateral ventricles. The figure follows the same color scheme as Fig.\ref{fig:hippocampus_modes}, the darker the color on the voxel is, the larger the deformation is for that particular region.}
		\label{fig:hcad}
	\end{figure}
	
	\subsection{Multi-Structure Generative Model}\label{sec:experiments_gen_mult}
	In our final experiment, we demonstrate the generative model for multiple brain structures. 
	We extend two of the experiments presented in the previous section to jointly modeling four structures: left and right hippocampi and left and right lateral ventricles. The split and the number of points per structure are the same as in the single-structure experiments (100/100 and 1024, respectively).
	
	\textbf{Reconstruction Error:}
	We evaluate the effect that conditioning and using multi-structure information have on obtaining accurate and compact representations of all the shapes in parallel. As in section \ref{sec:experiments_gen_single}, the reconstruction error is computed with EMD between original and generated point clouds, where we are averaging the EMD across multiple structures, as seen in Eq.(\ref{eq:emd_avg}).

	As we want to study the effect of conditioning and the effect of jointly modeling multiple structures,  we evaluate five different scenarios: 
	\begin{enumerate}
	    \item Generating each shape using Shapeworks
	    \item Training an independent non-conditional single structure generative model (Single Structure) for each structure
	    \item Training an independent conditional single structure generative model (Single Structure Conditional) for each structure (with the same one-hot encoding as in section \ref{sec:experiments_gen_single}).
	    \item Training non-conditional multi-structure generative model (Multi-structure) for all the structures.
	    \item Training conditional multi-structure generative model (Multi-structure Conditional) for all the structures (with the same one-hot encoding as in section \ref{sec:experiments_gen_single}).
	\end{enumerate}
	
	Fig.\ref{fig:reconstruction_error_curve} illustrates the averaged reconstruction error when changing the number of latent features used for encoding each structure. Notice that in case of the multi-structure models, a uniform distribution of the number of features per structure is assumed (we divide the dimension of the latent space by the number of structures that are used). We  observe that adding multiple structures and conditioning the latent vectors leads to the lowest reconstruction error per structure. As a matter of fact, if we compare the scenario where a four-feature vector is used for each structure (Single Structure Conditional) against the case where a four-feature (one per structure) vector is used to encode all the structures (Multi-structure Conditional), the reconstruction error is practically the same, showing that the multi-structural method provides a more efficient representation.

	\textbf{Effect of conditioning the shape model using a diagnostic label:} 
	As for the single structure experiment, we evaluate the effect of the condition vector in the model, but this time on all four structures. 
	Fig.~\ref{fig:hcad} shows synthetically generated shapes for healthy controls and AD patients, by setting the condition vector accordingly. 
	The difference between HC and AD shapes is illustrated in color. 
	We observe that the most critical areas in the hippocampus (dark red) are again placed in the lateral part of the hippocampus body, roughly around the CA1 subfield. 
	This is consistent to the previous results on the single structure analysis. 
	While we also observe differences for the lateral ventricles, they are not as pronounced as for the hippocampus. 

	\subsection{Summary of Experiments}
	Our results on Alzheimer's prediction and age regression have demonstrated the advantage of learning shape representations. Interestingly, even shape features that are not optimized for that particular task, i.e., generative features, achieve state-of-the-art results. 
	The advantage of working with the generative features is that we can reconstruct shapes from the low-dimensional embedding. 
	This helps in understanding the representations that have been learned and to perform shape manipulations in the low-dimensional space. 
	\revision{Other shape representations like spectral signatures do not offer the possibility for reconstructing the shape from arbitrary points in latent space.}  \revision{Moreover, the additional cost of classifying/processing shapes after training is negligible ($<20$ms per patient on an NVIDIA GPU GTX1080) compared to non-deep learning approaches.}
	
	Our results have further demonstrated the potential of the conditional vector. 
	In the MNIST experiment, we have trained a single model for all numbers and by setting the condition accordingly, we were able to reconstruct specific numbers, where the latent space among all numbers is shared. 
	The condition vector therefore presents a powerful means to include non-image information in the model. 
	Similar results for patients with Alzheimer's disease showed that shape changes associated to dementia can be identified. 
	Training a classifier on these synthetically generated shapes for AD classification was even able to obtain a higher accuracy than a directly trained model. 
	Finally, the extension to multiple structures leads to a more efficient encoding than working with multiple individual models, as relevant information can be shared. 
	
	\revision{We used the same number of points per point cloud within each experiment. 
	For the discriminative network, we could also work with varying numbers of points, since the GSN will output a feature vector with a fixed length. 
	For the generative model, in contrast, we need to fix the number of points for the decoder as it consists of an MLP. 
	In our experiments, we used up to 1,024 points per point cloud, which was sufficient to capture the geometry of subcortical structures and MNIST numbers. 
	For structures with more complex geometry, e.g., the cortex, a larger number of points would be necessary to have a faithful representation, which could be challenging for our framework as there are GPU memory constraints. 
	Yet, we would also like to emphasize that other shape approaches like ShapeWorks are more limited in this regard.}

	\section{Conclusions}
	
	We have presented a framework for anatomical shape analysis with deep neural networks and demonstrated its application for discriminative and generative tasks. 
	The framework consists of several computational blocks that cover a variety of shape processing tasks. 
	It takes unordered point clouds as input, without the need of correspondences between them, and extracts features that can be used for different applications, e.g., classification and reconstruction. 
	Next to the analysis of single structures, we have presented an extension that learns feature representations from multiple structures simultaneously. 
	In exhaustive experiments, we have demonstrated the advantages of the framework on real and synthetic data. 
	The discriminative model results in high accuracy in both classification and regression tasks, when compared to alternative shape descriptors.
	The generative model can encode complex shape variations using a low-dimensional embedding that leads to a low reconstruction error.
	Further, the introduction of a condition vector enables to integrate phenotypic information and additional control on the reconstruction. 
	We have demonstrated the properties of  our generative model  by  creating realistically looking synthetic shapes, which can even be used to train deep learning based  models. This has the potential to enable the use of powerful models in scenarios where the amount of annotated data is limited.
	Finally, we have demonstrated the advantages of the joint modeling of multiple structures. 
	
	Overall, our network facilitates processing of large datasets, since we do not require expensive operations for finding point correspondences between samples.
	On the neuroanatomical experiments, we operated on relatively small sample sizes to ensure a fair comparison to previous approaches, but on the MNIST data we demonstrated that our network scales to datasets with thousands of shapes.
	We believe that our framework can also be used to analyze other anatomical structures and  more diverse  phenotypic information in the condition vector.

	\section*{Acknowledgements}
	This research was partially supported by DFG, BMBF and the Bavarian State Ministry of Science and the Arts and coordinated by the Bavarian Research Institute for Digital Transformation (bidt).

	Data collection and sharing for this project was funded by the Alzheimer's Disease
	Neuroimaging Initiative (ADNI) (National Institutes of Health Grant U01 AG024904) and
	DOD ADNI (Department of Defense award number W81XWH-12-2-0012). ADNI is funded
	by the National Institute on Aging, the National Institute of Biomedical Imaging and
	Bioengineering, and through generous contributions from the following: Alzheimer's
	Association; Alzheimer's Drug Discovery Foundation; Araclon Biotech; BioClinica, Inc.;
	Biogen Idec Inc.; Bristol-Myers Squibb Company; Eisai Inc.; Elan Pharmaceuticals, Inc.; Eli
	Lilly and Company; EuroImmun; F. Hoffmann-La Roche Ltd and its affiliated company
	Genentech, Inc.; Fujirebio; GE Healthcare; ; IXICO Ltd.; Janssen Alzheimer Immunotherapy
	Research \& Development, LLC.; Johnson \& Johnson Pharmaceutical Research \&
	Development LLC.; Medpace, Inc.; Merck \& Co., Inc.; Meso Scale Diagnostics,
	LLC.; NeuroRx Research; Neurotrack Technologies; Novartis Pharmaceuticals
	Corporation; Pfizer Inc.; Piramal Imaging; Servier; Synarc Inc.; and Takeda Pharmaceutical
	Company. The Canadian Institutes of Health Research is providing funds to support ADNI
	clinical sites in Canada. Private sector contributions are facilitated by the Foundation for the
	National Institutes of Health (www.fnih.org). The grantee organization is the Northern
	California Institute for Research and Education, and the study is coordinated by the
	Alzheimer's Disease Cooperative Study at the University of California, San Diego. ADNI
	data are disseminated by the Laboratory for Neuro Imaging at the University of Southern
	California.

	\bibliographystyle{elsarticle-harv}
	\bibliography{biblio}

\begin{thebibliography}{40}
\expandafter\ifx\csname natexlab\endcsname\relax\def\natexlab#1{#1}\fi
\expandafter\ifx\csname url\endcsname\relax
  \def\url#1{\texttt{#1}}\fi
\expandafter\ifx\csname urlprefix\endcsname\relax\def\urlprefix{URL }\fi

\bibitem[{Becker et~al.(2018)Becker, Klein, and Wachinger}]{becker2018gaussian}
Becker, B.~G., Klein, T., Wachinger, C., 2018. Gaussian process uncertainty in
  age estimation as a measure of brain abnormality. NeuroImage.

\bibitem[{Bronstein et~al.(2017)Bronstein, Bruna, LeCun, Szlam, and
  Vandergheynst}]{Bronstein2017}
Bronstein, M.~M., Bruna, J., LeCun, Y., Szlam, A., Vandergheynst, P., 2017.
  Geometric deep learning: going beyond euclidean data. IEEE Signal Processing
  Magazine 34~(4), 18--42.

\bibitem[{Cates et~al.(2008)Cates, Fletcher, Styner, Hazlett, and
  Whitaker}]{cates2008particle}
Cates, J., Fletcher, P.~T., Styner, M., Hazlett, H.~C., Whitaker, R., 2008.
  Particle-based shape analysis of multi-object complexes. In: International
  Conference on Medical Image Computing and Computer-Assisted Intervention.
  Springer, pp. 477--485.

\bibitem[{Cerrolaza et~al.(2018)Cerrolaza, Li, Biffi, Gomez, Sinclair, Matthew,
  Knight, Kainz, and Rueckert}]{cerrolaza20183d}
Cerrolaza, J.~J., Li, Y., Biffi, C., Gomez, A., Sinclair, M., Matthew, J.,
  Knight, C., Kainz, B., Rueckert, D., 2018. 3d fetal skull reconstruction from
  2dus via deep conditional generative networks. In: International Conference
  on Medical Image Computing and Computer-Assisted Intervention. Springer, pp.
  383--391.

\bibitem[{Cootes et~al.(1995)Cootes, Taylor, Cooper, and Graham}]{Cootes1995}
Cootes, T., Taylor, C., Cooper, D., Graham, J., jan 1995. {Active Shape
  Models-Their Training and Application}. Comput. Vis. Image Underst. 61~(1),
  38--59.
\newline\urlprefix\url{http://www.sciencedirect.com/science/article/pii/S1077314285710041}

\bibitem[{Cootes and Taylor(1992)}]{cootes1992active}
Cootes, T.~F., Taylor, C.~J., 1992. Active shape models—‘smart snakes’.
  In: BMVC92. Springer, pp. 266--275.

\bibitem[{Costafreda et~al.(2011)Costafreda, Dinov, Tu, Shi, Liu, Kloszewska,
  Mecocci, Soininen, Tsolaki, Vellas, et~al.}]{costafreda2011automated}
Costafreda, S.~G., Dinov, I.~D., Tu, Z., Shi, Y., Liu, C.-Y., Kloszewska, I.,
  Mecocci, P., Soininen, H., Tsolaki, M., Vellas, B., et~al., 2011. Automated
  hippocampal shape analysis predicts the onset of dementia in mild cognitive
  impairment. Neuroimage 56~(1), 212--219.

\bibitem[{Durrleman et~al.(2014)Durrleman, Prastawa, Charon, Korenberg, Joshi,
  Gerig, and Trouv{\'e}}]{durrleman2014morphometry}
Durrleman, S., Prastawa, M., Charon, N., Korenberg, J.~R., Joshi, S., Gerig,
  G., Trouv{\'e}, A., 2014. Morphometry of anatomical shape complexes with
  dense deformations and sparse parameters. NeuroImage 101, 35--49.

\bibitem[{Evan and Sabuncu(2019)}]{evan2019convolutional}
Evan, M.~Y., Sabuncu, M.~R., 2019. A convolutional autoencoder approach to
  learn volumetric shape representations for brain structures. In: 2019 IEEE
  16th International Symposium on Biomedical Imaging (ISBI 2019). IEEE, pp.
  1559--1562.

\bibitem[{Fischl(2012)}]{Fischl2012}
Fischl, B., 2012. Freesurfer. Neuroimage 62~(2), 774--781.

\bibitem[{Franke et~al.(2010)Franke, Ziegler, Kl{\"o}ppel, Gaser, Initiative,
  et~al.}]{franke2010estimating}
Franke, K., Ziegler, G., Kl{\"o}ppel, S., Gaser, C., Initiative, A. D.~N.,
  et~al., 2010. Estimating the age of healthy subjects from t1-weighted mri
  scans using kernel methods: exploring the influence of various parameters.
  Neuroimage 50~(3), 883--892.

\bibitem[{Frisoni et~al.(2008)Frisoni, Ganzola, Canu, R{\"u}b, Pizzini,
  Alessandrini, Zoccatelli, Beltramello, Caltagirone, and
  Thompson}]{frisoni2008mapping}
Frisoni, G.~B., Ganzola, R., Canu, E., R{\"u}b, U., Pizzini, F.~B.,
  Alessandrini, F., Zoccatelli, G., Beltramello, A., Caltagirone, C., Thompson,
  P.~M., 2008. Mapping local hippocampal changes in alzheimer's disease and
  normal ageing with mri at 3 tesla. Brain 131~(12), 3266--3276.

\bibitem[{Gerardin et~al.(2009)Gerardin, Ch{\'e}telat, Chupin, Cuingnet,
  Desgranges, Kim, Niethammer, Dubois, Leh{\'e}ricy, Garnero,
  et~al.}]{gerardin2009multidimensional}
Gerardin, E., Ch{\'e}telat, G., Chupin, M., Cuingnet, R., Desgranges, B., Kim,
  H.-S., Niethammer, M., Dubois, B., Leh{\'e}ricy, S., Garnero, L., et~al.,
  2009. Multidimensional classification of hippocampal shape features
  discriminates alzheimer's disease and mild cognitive impairment from normal
  aging. Neuroimage 47~(4), 1476--1486.

\bibitem[{Goodfellow et~al.(2014)Goodfellow, Pouget-Abadie, Mirza, Xu,
  Warde-Farley, Ozair, Courville, and Bengio}]{goodfellow2014generative}
Goodfellow, I., Pouget-Abadie, J., Mirza, M., Xu, B., Warde-Farley, D., Ozair,
  S., Courville, A., Bengio, Y., 2014. Generative adversarial nets. In:
  Advances in neural information processing systems. pp. 2672--2680.

\bibitem[{Goparaju et~al.(2018)Goparaju, Csecs, Morris, Kholmovski, Marrouche,
  Whitaker, and Elhabian}]{goparaju2018evaluation}
Goparaju, A., Csecs, I., Morris, A., Kholmovski, E., Marrouche, N., Whitaker,
  R., Elhabian, S., 2018. On the evaluation and validation of off-the-shelf
  statistical shape modeling tools: A clinical application. In: International
  Workshop on Shape in Medical Imaging. Springer, pp. 14--27.

\bibitem[{Gorczowski et~al.(2007)Gorczowski, Styner, Jeong, Marron, Piven,
  Hazlett, Pizer, and Gerig}]{Gorczowski2007}
Gorczowski, K., Styner, M., Jeong, J.-Y., Marron, J., Piven, J., Hazlett,
  H.~C., Pizer, S.~M., Gerig, G., 2007. Statistical shape analysis of
  multi-object complexes. In: Computer Vision and Pattern Recognition, 2007.
  CVPR'07. IEEE Conference on. IEEE, pp. 1--8.

\bibitem[{Guti{\'e}rrez-Becker and Wachinger(2018)}]{gutierrez2018deep}
Guti{\'e}rrez-Becker, B., Wachinger, C., 2018. Deep multi-structural shape
  analysis: Application to neuroanatomy. International Conference on Medical
  Image Computing and Computer-Assisted Intervention.

\bibitem[{Guti{\'e}rrez-Becker and Wachinger(2019)}]{gutierrez2019learning}
Guti{\'e}rrez-Becker, B., Wachinger, C., 2019. Learning a conditional
  generative model for anatomical shape analysis. In: International Conference
  on Information Processing in Medical Imaging. Springer, pp. 505--516.

\bibitem[{He et~al.(2019)He, Gopinath, Desrosiers, and
  Lombaert}]{he2019spectral}
He, R., Gopinath, K., Desrosiers, C., Lombaert, H., 2019. Spectral graph
  transformer networks for brain surface parcellation. arXiv preprint
  arXiv:1911.10118.

\bibitem[{Isola et~al.(2017)Isola, Zhu, Zhou, and Efros}]{isola2017image}
Isola, P., Zhu, J.-Y., Zhou, T., Efros, A.~A., 2017. Image-to-image translation
  with conditional adversarial networks. In: 2017 IEEE Conference on Computer
  Vision and Pattern Recognition (CVPR). IEEE, pp. 5967--5976.

\bibitem[{Jack et~al.(2008)Jack, Bernstein, Fox, Thompson, Alexander, Harvey,
  Borowski, Britson, L~Whitwell, Ward, et~al.}]{Jack2008}
Jack, C.~R., Bernstein, M.~A., Fox, N.~C., Thompson, P., Alexander, G., Harvey,
  D., Borowski, B., Britson, P.~J., L~Whitwell, J., Ward, C., et~al., 2008. The
  alzheimer's disease neuroimaging initiative (adni): Mri methods. Journal of
  magnetic resonance imaging 27~(4), 685--691.

\bibitem[{Jaderberg et~al.(2015)Jaderberg, Simonyan, Zisserman,
  et~al.}]{jaderberg2015spatial}
Jaderberg, M., Simonyan, K., Zisserman, A., et~al., 2015. Spatial transformer
  networks. In: Advances in neural information processing systems. pp.
  2017--2025.

\bibitem[{Kendall(1989)}]{kendall1989survey}
Kendall, D.~G., 1989. A survey of the statistical theory of shape. Statistical
  Science, 87--99.

\bibitem[{Kingma and Welling(2013)}]{kingma2013auto}
Kingma, D.~P., Welling, M., 2013. Auto-encoding variational bayes. arXiv
  preprint arXiv:1312.6114.

\bibitem[{Litjens et~al.(2017)Litjens, Kooi, Bejnordi, Setio, Ciompi,
  Ghafoorian, Van Der~Laak, Van~Ginneken, and S{\'a}nchez}]{litjens2017survey}
Litjens, G., Kooi, T., Bejnordi, B.~E., Setio, A. A.~A., Ciompi, F.,
  Ghafoorian, M., Van Der~Laak, J.~A., Van~Ginneken, B., S{\'a}nchez, C.~I.,
  2017. A survey on deep learning in medical image analysis. Medical image
  analysis 42, 60--88.

\bibitem[{Lombaert et~al.(2013)Lombaert, Sporring, and
  Siddiqi}]{lombaert2013diffeomorphic}
Lombaert, H., Sporring, J., Siddiqi, K., 2013. Diffeomorphic spectral matching
  of cortical surfaces. In: International Conference on Information Processing
  in Medical Imaging. Springer, pp. 376--389.

\bibitem[{Miller et~al.(2014)Miller, Younes, and
  Trouv{\'e}}]{miller2014diffeomorphometry}
Miller, M.~I., Younes, L., Trouv{\'e}, A., 2014. Diffeomorphometry and geodesic
  positioning systems for human anatomy. Technology 2~(01), 36--43.

\bibitem[{Ng et~al.(2014)Ng, Toews, Durrleman, and Shi}]{ng2014shape}
Ng, B., Toews, M., Durrleman, S., Shi, Y., 2014. Shape analysis for brain
  structures. In: Shape Analysis in Medical Image Analysis. Springer, pp.
  3--49.

\bibitem[{Pennec et~al.(2019)Pennec, Sommer, and
  Fletcher}]{pennec2019riemannian}
Pennec, X., Sommer, S., Fletcher, T., 2019. Riemannian Geometric Statistics in
  Medical Image Analysis. Academic Press.

\bibitem[{Pizer et~al.(2013)Pizer, Jung, Goswami, Vicory, Zhao, Chaudhuri,
  Damon, Huckemann, and Marron}]{pizer2013nested}
Pizer, S.~M., Jung, S., Goswami, D., Vicory, J., Zhao, X., Chaudhuri, R.,
  Damon, J.~N., Huckemann, S., Marron, J., 2013. Nested sphere statistics of
  skeletal models. In: Innovations for Shape Analysis. Springer, pp. 93--115.

\bibitem[{Qi et~al.(2017)Qi, Su, Mo, and Guibas}]{Qi2017}
Qi, C.~R., Su, H., Mo, K., Guibas, L.~J., 2017. Pointnet: Deep learning on
  point sets for 3d classification and segmentation. Proc. Computer Vision and
  Pattern Recognition (CVPR), IEEE 1~(2), 4.

\bibitem[{Ranjan et~al.(2018)Ranjan, Bolkart, Sanyal, and
  Black}]{ranjan2018generating}
Ranjan, A., Bolkart, T., Sanyal, S., Black, M.~J., 2018. Generating 3d faces
  using convolutional mesh autoencoders. In: Proceedings of the European
  Conference on Computer Vision (ECCV). pp. 704--720.

\bibitem[{Reuter et~al.(2006)Reuter, Wolter, and Peinecke}]{reuter2006laplace}
Reuter, M., Wolter, F.-E., Peinecke, N., 2006. Laplace--beltrami spectra as
  ‘shape-dna’of surfaces and solids. Computer-Aided Design 38~(4),
  342--366.

\bibitem[{Rubner et~al.(2000)Rubner, Tomasi, and Guibas}]{rubner2000earth}
Rubner, Y., Tomasi, C., Guibas, L.~J., 2000. The earth mover's distance as a
  metric for image retrieval. International journal of computer vision 40~(2),
  99--121.

\bibitem[{Shakeri et~al.(2016)Shakeri, Lombaert, Tripathi, Kadoury, Initiative,
  et~al.}]{Shakeri2016}
Shakeri, M., Lombaert, H., Tripathi, S., Kadoury, S., Initiative, A. D.~N.,
  et~al., 2016. Deep spectral-based shape features for alzheimer's disease
  classification. In: International Workshop on Spectral and Shape Analysis in
  Medical Imaging. Springer, pp. 15--24.

\bibitem[{Shen et~al.(2012)Shen, Fripp, M{\'e}riaudeau, Ch{\'e}telat, Salvado,
  and Bourgeat}]{shen2012detecting}
Shen, K.-k., Fripp, J., M{\'e}riaudeau, F., Ch{\'e}telat, G., Salvado, O.,
  Bourgeat, P., 2012. Detecting global and local hippocampal shape changes in
  alzheimer's disease using statistical shape models. Neuroimage 59~(3),
  2155--2166.

\bibitem[{Sohn et~al.(2015)Sohn, Lee, and Yan}]{Sohn2015}
Sohn, K., Lee, H., Yan, X., 2015. Learning structured output representation
  using deep conditional generative models. In: Advances in Neural Information
  Processing Systems. pp. 3483--3491.

\bibitem[{Wachinger et~al.(2015)Wachinger, Golland, Kremen, Fischl, and
  Reuter}]{Wachinger2015}
Wachinger, C., Golland, P., Kremen, W., Fischl, B., Reuter, M., 2015.
  {BrainPrint: A discriminative characterization of brain morphology}.
  Neuroimage 109, 232--248.

\bibitem[{Wachinger and Reuter(2016)}]{wachinger2016domain}
Wachinger, C., Reuter, M., 2016. Domain adaptation for alzheimer's disease
  diagnostics. Neuroimage 139, 470--479.

\bibitem[{Wachinger et~al.(2017)Wachinger, Rieckmann, and
  Reuter}]{wachinger2017latent}
Wachinger, C., Rieckmann, A., Reuter, M., 2017. Latent processes governing
  neuroanatomical change in aging and dementia. In: International Conference on
  Medical Image Computing and Computer-Assisted Intervention. Springer, pp.
  30--37.

\end{thebibliography}
\end{document}